\documentclass[final,twocolumn,5p,authoryear]{elsarticle}

\usepackage[sort,compress]{cite}
\usepackage{graphicx} % for pdf, bitmapped graphics files
\usepackage{subfigure}
\usepackage{optidef} % For optimization problem writing
\usepackage{amsmath} % assumes amsmath package installed
\usepackage{amssymb}  % assumes amsmath package installed
\usepackage{color}
\usepackage[ruled]{algorithm2e}
\usepackage{algorithmic}
\usepackage{enumitem}
\usepackage{booktabs}
\usepackage{hyperref} % For href command in emails
\usepackage{amsthm}
\theoremstyle{definition}
\journal{Robotics and Autonomous Systems}

\newcommand{\blu}[1]{{\color{black} #1}}

\usepackage[normalem]{ulem} % Ulem package to get strike through font

\title{\LARGE \bf
A Survey of Multi-Agent Human-Robot Interaction Systems}

\begin{document}
\begin{frontmatter}

\author[inst1]{Abhinav Dahiya\corref{cor1}}
\author[inst1]{Alexander M. Aroyo}
\author[inst1,inst2]{Kerstin Dautenhahn}
\author[inst1]{Stephen L. Smith}

\affiliation[inst1]{organization={Department of Electrical and Computer Engineering},%Department and Organization
            addressline={University of Waterloo, Waterloo, ON}, 
            country={Canada}}
\affiliation[inst2]{organization={Department of Systems Design Engineering},%Department and Organization
            addressline={University of Waterloo, Waterloo, ON}, 
            country={Canada}}

\cortext[cor1]{Corresponding author. \\Email address: abhinav.dahiya@uwaterloo.ca}
\tnotetext[]{This research is supported in part by the Natural Sciences and Engineering Research Council of Canada (NSERC), in part by the Innovation for Defence Excellence and Security (IDEaS) Program of the Canadian Department of National Defence through grant CFPMN2-037, and in part by funding from the Canada 150 Research Chairs Program.}

\begin{abstract}
This article presents a survey of literature in the area of Human-Robot Interaction (HRI), specifically on systems containing more than two agents (i.e., having multiple humans and/or multiple robots).  We identify three core aspects of ``Multi-agent" HRI systems that are useful for understanding how these systems differ from dyadic systems and from one another.  These are the Team structure, Interaction style among agents, and the system's Computational characteristics.
Under these core aspects, we present five attributes of HRI systems, namely Team size, Team composition, Interaction model, Communication modalities, and Robot control.  These attributes are used to characterize and distinguish one system from another.  
We populate resulting categories with examples from the recent literature along with a brief discussion of their applications. We also analyze how these attributes in multi-agent systems differ from the case of dyadic human-robot systems.  
Through this survey, we summarize key observations from the current literature, and identify challenges and promising areas for future research in this domain.  
In order to realize the vision of robots being part of the society and interacting seamlessly with humans, there is a need to expand research on multi-human -- multi-robot systems.  Not only do these systems require coordination among several agents, they also involve multi-agent and indirect interactions which are absent from dyadic HRI systems.  Adding multiple agents in HRI systems requires more advanced interaction schemes, behavior understanding and control methods to allow natural interactions among humans and robots.  
In addition, research on human behavioral understanding in mixed human-robot teams also requires more attention.  This will help formulate and implement effective robot control policies in HRI systems with large numbers of heterogeneous robots and humans; a team composition reflecting many real-world scenarios.
\end{abstract}

\begin{highlights}
\item Multi-agent systems in the Human-Robot Interaction (HRI) literature are surveyed
\item Graph representation of interaction models is introduced for multi-agent HRI systems
\item Three core aspects of multi-agent HRI systems are identified: Team structure, Interaction style, and Computational characteristics
\item Under these core aspects, HRI systems discussed under five attributes like Team composition, Interaction model etc. %including Team size, Team composition, Interaction model etc.
\item The survey points out that multi-human -- multi-robot systems need more attention
\item More research is also required on human behavioral understanding in mixed/heterogeneous human-robot teams % also requires more attention to better understand mixed/heterogeneous human-robot teams

\end{highlights}


\end{frontmatter}

%%%%%%%%%%%%%%%%%%%%%%%%%%%%%%%%%%%%%%%%%%%%%%%%%%%%%%%%%%%%%%%%%%%%%%%%%%%%%%%%
\section{Introduction}
% %
\begin{figure*}[ht]
    \centering
    \includegraphics[width=1.5\columnwidth]{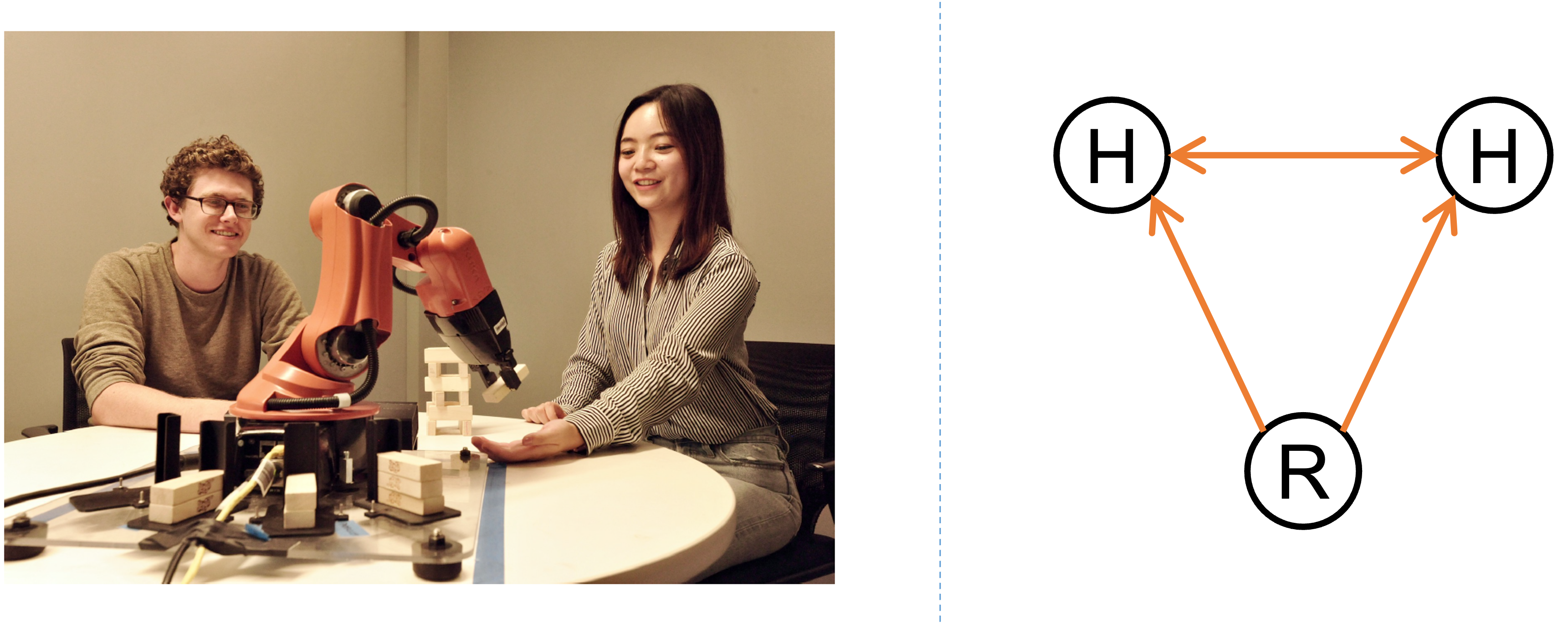} \caption{Left: HRI system for robot-assisted tower construction task where the robot allocates tower pieces between two human teammates \citep{jung2020robot}. Right: The corresponding interaction graph representation of the system.  In the graph, each human and robot is denoted by a node.  Each arrow represents an interaction and its direction signifies the direction of information flow.  When information flow between two nodes is present in both directions, a bi-directional arrow is used.  In this system, human behavior is dependent on robot's actions while the robot's actions are independent of its human teammates (thus the uni-directional arrows).  Both humans can potentially discuss their next actions with the other human teammate (denoted by the bi-directional arrow).}
    \label{fig:interaction_arrows}
\end{figure*}

Research in Human-Robot Interaction (HRI) has facilitated the introduction of robots in human spaces, and has gained significant momentum in the past two decades.  Existing literature indicate that the vast majority of  past research pertains to \textit{dyadic} systems where a single human interacts with a single robot.
However, this trend is changing now.  With the improvement in physical and computational capabilities of robotic systems, we see robots working in teams of more than two, moving beyond the typical dyadic HRI systems.
\blu{Even though multi-agent HRI systems are built on similar infrastructure as dyadic systems, there are several differences.  Multi-agent HRI systems require more complex control strategies to coordinate several agents that may be dissimilar to one another (in roles, communication modes etc.), can involve interactions that connect more than two agents at once, and need special attention to address any conflicts that may arise from their interactions.  These differences mean that control strategies developed for dyadic systems are either not applicable to multi-agent systems or they do not work as intended.  Therefore, in the research literature, we see numerous studies addressing one or more aspects of multi-agent HRI systems specifically.}
Through this survey we aim to organize the existing work, and suggest  a structure through which researchers can compare and contrast the existing and future research.

A human-robot system can have a varied number of agents, both humans and robots, interacting with each other.  Even though any human-robot system is technically a multi-agent system (comprised of at least one human and one robot), in the HRI literature, the terminology is used differently.  In HRI literature, \textit{dyadic} systems refer to the ones built around interactions between exactly one robot and one human.  \textit{Triadic} systems have three agents, with either two humans and a robot, or two robots and a human \citep{gvirsman2020patricc, gombolay2015decision}.  For systems with three or more agents, terms such as \textit{teams} or \textit{groups} are used \citep{abrams2020c}.  In this article, we use the term \textit{multi-agent} to refer to any human-robot system containing more than one human (multi-human) and/or more than one robot (multi-robot).  
\blu{Robots assisting humans in an industrial task, a group of humans and robots doing a social activity, and human operators controlling multiple robots are some examples of multi-agent HRI systems.  Such systems are shown in Figures~\ref{fig:team_comp_images}(b), \ref{fig:interaction_models_images}(b) and \ref{fig:interaction_models_images}(c) respectively.  This article frequently refers to related work on different multi-agent HRI systems and illustrates the work in the subsequent sections.  Figure~\ref{fig:interaction_arrows}, as an example of a multi-agent HRI system, shows a system with two humans being assisted by a robot arm in a tower construction task.}

In the literature, we find several studies and articles that review  research in the field of HRI and Collaboration, e.g.  \citep{selvaggio2021autonomy, bartneck2020human, sheridan2016human, thomaz2016computational}.  
There are also studies presenting taxonomies and classification criteria for HRI systems in general, and reviewing the work done in different social and industrial applications of HRI, e.g. \citep{wang2017human, yanco2004classifying}.  
\citet{lewis2013human} reviews the literature on human supervision of multiple robots that are situated remotely, and analyses such systems under the notion of \textit{command complexity}. % and examines its implications for the users.
\citet{thomaz2016computational} present a detailed review of the computational aspects of HRI studies, covering topics such as algorithms, modelling and computational framework design.
Human-Swarm Interaction (HSI) is another subset of multi-agent systems where a small number of humans (commonly one) interact with a group of robots that coordinate among themselves and often act as an unified entity.  \citet{kolling2015human} present a comprehensive survey of Human-Swarm interaction systems and discuss core concepts for their design.
In the multi-agent HRI literature, \citet{sebo2020robots} present a review of studies involving robots' interactions with a group or team of people, and explores characteristics of such systems.  The article discusses the role a robot plays in a group of humans and the influence it has in such groups. \blu{The main focus of their article is on human-robot systems consisting of multiple humans irrespective of the number of robots.  In \citet{oliveira2021human}, the authors discuss methodological and transversal issues in research on HRI in small groups from the perspective of social psychology.}

{To extend the existing multi-agent HRI literature, in this paper, we expand the scope of existing surveys and reviews discussed above by including systems having multiple robots interacting with either a single human or a group of humans, in and out of a social setting.  We also consider systems built around interactions among multiple humans and multiple robots simultaneously.  In addition, we also include studies on the theoretical/computational research, where actual robots or humans are not present, but which still contain crucial aspects of an HRI system (e.g., scheduling when to collaborate \citep{hari2020approximation, ijtsma2019computational}, optimizing collaborative manipulation \citep{liu2021optimized} etc.). These systems are a valuable part of multi-agent HRI literature and provide useful insight into implementation of real systems.}

Research in Human-Robot Interaction has a wide scope, and includes various aspects of robot decision-making, human behavioral modelling and system analysis.  This research area includes any system where robot(s) and human(s) co-exist, in or out of a team setting, and working towards shared or distinct objectives.  This co-existence can either be in the same physical environment, resulting in co-located interactions, or it can emerge in a system where remotely present agents are connected through a \textit{virtual} interface.  In addition, work on developing capable and efficient algorithms to improve interaction between humans and robots also has an important place in HRI research.  
There are three kinds of studies (or a combination of these) present in the research literature: 1) System Design: ones that propose a system useful in the context of HRI (direct application; \blu{e.g., \citet{karami2020task}}), 2) Observational: ones that detail a study designed to elicit different features of HRI systems (to understand system behavior; \blu{e.g., \citet{podevijn2016investigating}}) and 3) Algorithmic Design: ones that present computational research, including algorithms, models and frameworks to solve the decision-making problems in HRI systems \blu{(e.g., \citet{dahiya2022scalable})}.  In this survey we include all three kinds of studies from the multi-agent HRI literature.

\textbf{Our inclusion/exclusion criteria:} Papers included in this survey article, different from systematic reviews, were found using online search through issues and proceedings of various journals and conferences related to the HRI research.  The search was further expanded using the citations of the papers found in the first step.  To be included, the research work must have an interaction aspect (either real or synthetic interactions) between at least three agents \footnote{This article does include a few studies with dyadic systems in order to discuss relevant details.} with at least one human and one robot, and the intended application should be in human-robot interaction.  This means we excluded some human-agent interaction studies which may be, for example, psychology-oriented.  We also exclude studies where humans are used merely for data collection, or where the agent is merely a computational entity without any direct application to the human-robot interaction aspect, e.g., studies on learning from demonstration, or intelligent recommendation systems.

In this survey, we identify five attributes through which multi-agent HRI systems can be characterized and compared with each other.  These attributes stem from three core aspects of multi-agent HRI systems: the team structure/architecture, interaction style, and computational characteristics (Section~\ref{sec:general_classification}).  Subsequent sections introduce the resulting categorizations in detail and provide examples of human-robot interaction systems from the current literature (Sections~\ref{sec:team_structure}-\ref{sec:control}).  
% This categorization has enabled us to identify various patterns in the literature and correlations that exist between different features of such systems.  
Several images of multi-agent HRI systems from the literature are also shown to illustrate different categories\footnote{It is important to note that one can only show images of systems where it is possible to capture all agents in a single frame. However, the scope of multi-agent HRI systems goes far beyond what can be shown in figures, e.g., systems where interactions are separated spatially or temporally.}\footnote{Images used in the figures of this article are used with permission from respective copyright holders (publishers/authors).}.  Key observations regarding each categorization are listed at the end of the corresponding section.  
\blu{In Section~\ref{sec:interationsCoreAspects}, we comment on interactions between different core aspects and discuss what kind of systems are commonly seen in the literature, and how that may change in the future.}
The survey concludes with a discussion on main challenges currently being faced in multi-agent HRI research on various fronts and our insight on possible directions of future research in this field (Section~\ref{sec:directions}).

%%%%%%%%%%%%%%%%%%%%%%%%%%%%%

\section{Multi-Agent Human-Robot Interaction}
\label{sec:general_classification}
\begin{figure*}[ht]
    \centering
    \includegraphics[width=0.95\textwidth]{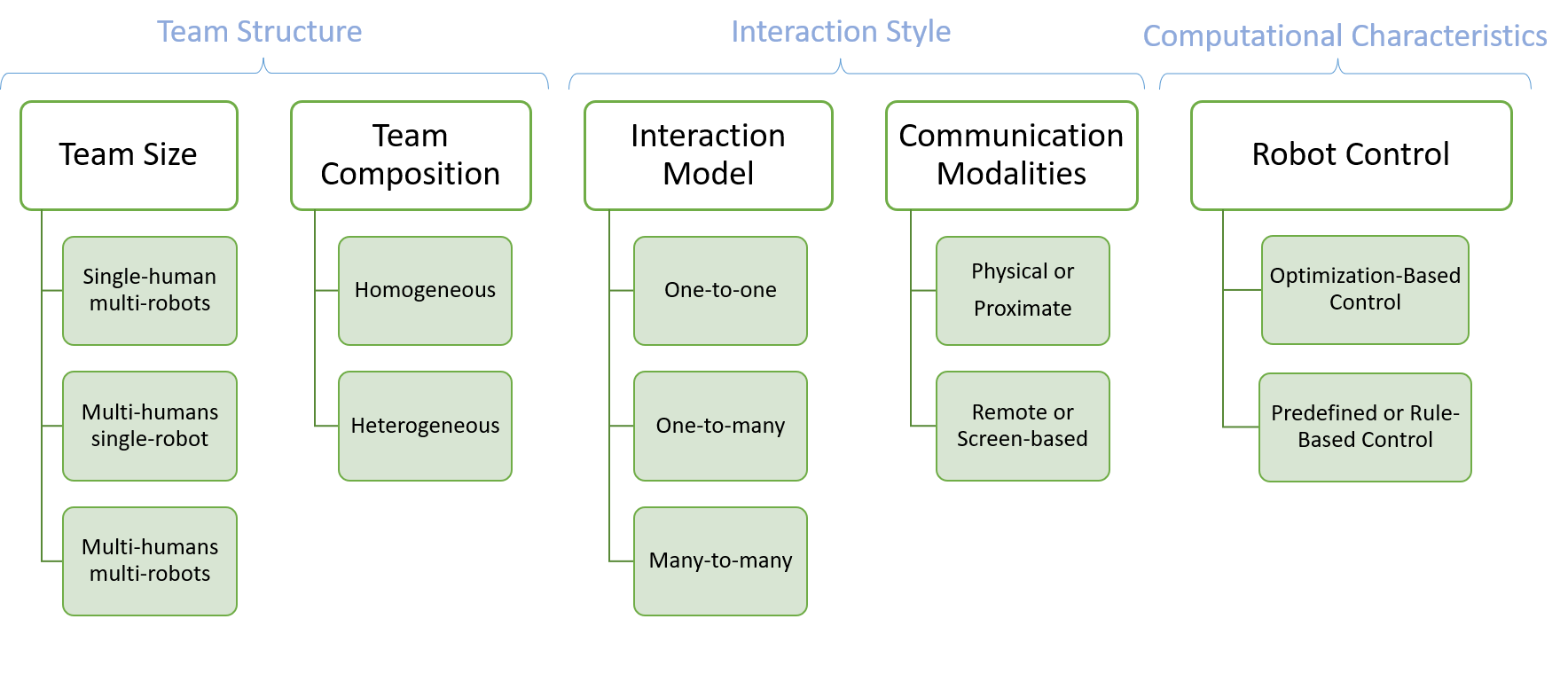}
    \caption{The different categorizations for multi-agent Human-Robot Interactions originating from the three core aspects: the team structure, the interaction style and the computational characteristics.  These aspects bring forth several attributes of systems, each with its own features and applications.}
    \label{fig:class_2}
\end{figure*}

In order to formally define multi-agent Human-Robot Interaction systems, we first need to define `interaction'. 
\blu{The American Psychological Association \citep{APA} defines a social interaction as} a process that involves reciprocal stimulation or response between two or more individuals. In the context of multi-agent HRI, however, reciprocity is not a necessary requirement and an interaction can be defined as: \textit{information flow between two or more agents occurring as a result of communication, action or mere presence of any of those agents}. 
This information flow results in change in actions, behavior or mental state of the agents on the receiving end.  
Using this definition, we can express a multi-agent HRI system using an interaction graph (Fig.~\ref{fig:interaction_arrows}).  The figure shows a multi-agent HRI system and the corresponding interaction graph.  Each agent is depicted as a node in a directed graph in which each edge represents interaction between two agents and points towards the direction of information flow. We use this interaction graph structure to define multi-agent HRI systems:

\begin{quote}
Multi-agent Human-Robot Interaction systems are the ones for which the interaction graph contains three or more nodes (agents), with at least three nodes connected via interaction edges, including at least one interaction edge between a human node and a robot node.
\end{quote}

The information exchange can happen by means of verbal/non-verbal communication or through an interaction interface.  The agents may be homogeneous in their roles and identities, or they can be heterogeneous and contribute differently in the system.  Moreover, these agents can be located in separate environments interacting over virtual channels or they can share same physical space.  In this survey, we consider the literature on multi-agent HRI systems and try to characterize them based on different properties.

Interactions in multi-agent human-robot systems do bear similarities with interactions in human-only groups.  As we discuss in later sections, several studies in the literature try to apply understanding of human-only groups to multi-agent human-robot systems (e.g., by using concepts of human group interactions from social sciences).  
When compared to robot-only groups, interactions in human-robot systems come with a higher uncertainty of outcome, and studies emphasize making human response to an interaction more predictable (e.g. by increasing human's trust in robots or by making robots' intentions clearer). 
In this paper, we also discuss how research from human-only and robot-only groups can help multi-agent HRI, and open new avenues for future research. 

Human-Robot Interactions among multiple agents, when compared to dyadic interactions, introduce a number of changes and complexities in the system.  With multiple agents present, multi-agent interactions and indirect interactions can emerge in the system.  It also becomes possible to have humans interacting with other humans and robots with other robots.  These Human-Human Interactions (HHI) and Robot-Robot Interactions (RRI) then ultimately affect (or are affected by) interactions between humans and robots.  The presence of multiple agents also require advanced communication modalities and interfaces to enable effective interaction among all agents.  As the number of agents grow in the system, modelling their behavior and controlling their actions becomes increasingly challenging.  We look at each of these aspects of multi-agent HRI systems separately.

We propose that multi-agent HRI systems can be characterized and compared to one another based on three core aspects: 1) Team structure, 2) Interaction style and 3) Computational characteristics.  These aspects represent the way a system is set up, and the way interactions take place and methods by which agents are controlled in that system. 

\textbf{Team Structure:} The first aspect relates to agents that constitute the human-robot system. % characterizes an HRI system using team-based parameters of the system.  
This includes parameters such as the number and types of agents (both humans and robots) present in the system, and homogeneity among different agents.  Homogeneity is a measure of similarities in roles, capabilities, embodiment and the authorities granted to each of the agents (deciding who can initiate an interaction, give commands, request assistance etc.).

\textbf{Interaction Style:} The second aspect of multi-agent HRI systems describes how the agents interact with each other. This includes factors such as modality of communication happening among the agents and interaction models being implemented.  This tells us the way in which interactions take place in the system (through a screen or speech etc.) and the agents involved (between two or among multiple agents).

\textbf{Computational Characteristics:} The third aspect looks into the software part of HRI systems and describes how the behavior of different agents is controlled or influenced.  This includes robot task planning algorithms, model-based/model-free controllers and other ways of deciding robots' actions in the system, and the ways in which they are used to influence human behavior.

These three aspects provide us a way to establish distinctions and comparisons among multi-agent HRI systems.
For instance, consider a system where a human operator is supervising a team of remote mobile robots (e.g., \citep{swamy2020scaled, rosenfeld2017intelligent}).  We can characterize this system as one with a single human interacting with multiple robots (team size), and one where all robots are similar to one another (homogeneity).  The human might be able to give high-level commands to the robot group or control them individually (interaction model), using a screen-based interface (communication modality). The robot control can be made adaptive to actions of the operator or designed to optimize some utility function (computational characteristics).

Under these core aspects, we can characterize HRI systems based on five different attributes as shown in Fig.~\ref{fig:class_2}.    
\blu{It should be noted that these core aspects are not independent of each other and a system attribute under one aspect can influence other system attributes.  This interaction is discussed in Section~\ref{sec:interationsCoreAspects}. 
Also note that these core aspects are not meant to be exhaustive or the only way of distinguishing multi-agent HRI systems.  The aspects are chosen such that they allow for a taxonomy that is both broad enough to be applicable to multi-agent HRI systems from different areas of research, and detailed enough to meaningfully characterize and compare those systems.} %, and summarize any trends that emerge relating different categories.
In the following sections, we consider the above three aspects and the associated attributes in more detail, and present the categorizations that arise under these attributes.  Under each category, we include examples from recent literature to understand its application, and analyze strengths and limitations of different types of systems.

\begin{figure*}[ht]
  \centering
  \subfigure[]{\includegraphics[height=3.6cm]{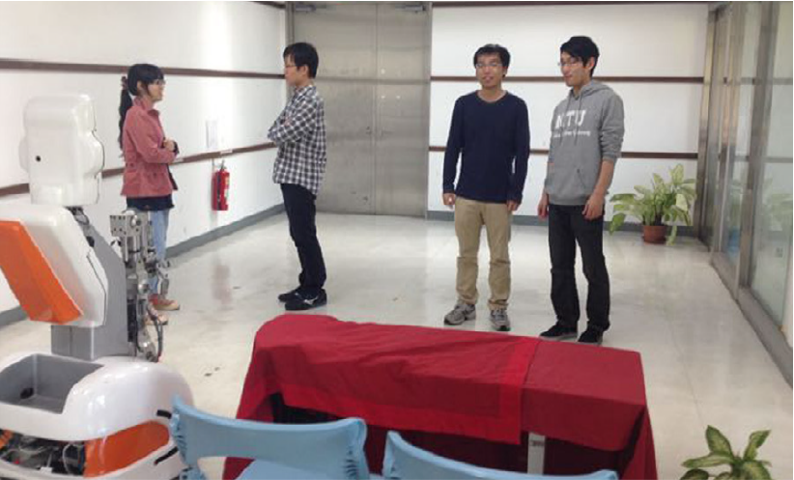}}\;
  \subfigure[]{\includegraphics[height=3.6cm]{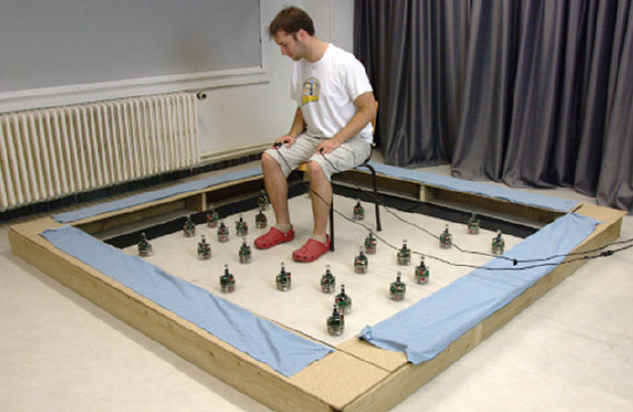}}\;
  \subfigure[]{\includegraphics[height=3.6cm]{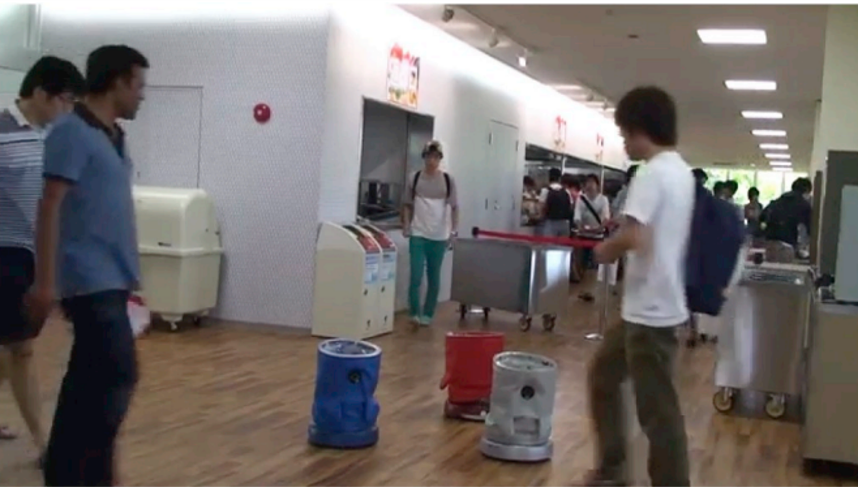}}
  \vspace{-0.75\baselineskip}
  \caption{Different possible configurations of multi-agent HRI systems based on team size.  a) A robot in a social environment \citep{tseng2016service};  b) Investigating effects of multiple robots on a human participant \citep{podevijn2016investigating};  c) Three robots interacting with each other and with the people around them \citep{fraune2020effects}.}
    \label{fig:team_size}
\end{figure*}

\section{Team Structure}
\label{sec:team_structure}
The most perceptible feature of a multi-agent HRI system is the size and composition of the human-robot team.  Depending on the application, a human-robot system can take advantage of more than one human and/or more than one robot in the team.  The task at hand may also require to utilize a team of agents with different capabilities and roles.
Both number and type of robots in a group can have significant effects on human's perceptions and emotions towards the robots \citep{fraune2015rabble}.
Since humans and robots usually have different ways of acting in a collaborative setting and interacting with their partners, the team structure decides various other aspects of the system including the way in which different agents interact, how they can share information and how their actions are planned \citep{fincannon2013best, fincannon2011team}.  
Based on team structure, multi-agent systems can differ from dyadic systems in two (quite apparent) ways: 
\begin{enumerate}
    \item There can be several possibilities of having different numbers of agents (both humans and robots) in the team, 
    \item There exists a notion of homogeneity/heterogeneity among the agents. 
\end{enumerate}

Therefore, we discuss in the following two factors under team structure: Team size, and Team composition.

%%%%%%%%%%%%%%%%%%%%%%%%%%%%%%%%%%%%%%%%%%%%%%%%%%%%%%%%%%%%%%%%%%%%%%%%
\subsection{Team Size}
Team size in an HRI system refers to the number of humans and robots present in the system.  Including single-human -- single-robot systems, the HRI systems can be grouped into the following categories based on their team size:

\subsubsection{Single-human -- single-robot} 
These are the conventional systems containing a single human interacting with a single robot and are the most common type of systems studied in HRI.  
%In relation to the multi-agent setting, it may be possible to extend such systems to a multi-agent one with slight modifications.  
This article, however, focuses mainly on systems with a higher number of agents and we refer the readers to other articles available in the literature for a review of single-human -- single-robot systems, and HRI in general \citep{bartneck2020human, breazeal2016social, goodrich2008human, bauer2008human}.
%%%%%%%%%%%%%%%%%%%%%%%%%%%%%%%%%%%%%%%%%%%%%%%%%%%%%%%%%%%%%%%%

\subsubsection{Single-human -- multi-robot} 
\label{sec:1hmR}
These systems comprise teams with a single human interacting or collaborating with multiple robots through the task.  % The interaction of the human with the robots may happen simultaneously or independently.
Such systems have found their utility in applications where the task is primarily executed by a number of (semi)autonomous robots requiring intermittent interventions/assistance from a human operator or supervisor, either in event of a \textit{fault} \citep{swamy2020scaled, wang2014human} or to further increase performance of the multi-robot team in areas like large-scale assembly \citep{sellner2006coordinated} and search-and-rescue \citep{khasawneh2019human, wang2008assessing}.  These systems are also employed in Human-Swarm Interaction \citep{kolling2015human} where multiple robots coordinate among themselves while receiving inputs from the human teammate.  
% This type of team composition is commonly seen in control of multiple robots for a variety of applications such as object retrieval \citep{frank2017toward}, search-and-rescue \citep{ruff2002human, guo2009touch, shiomi2009field, crandall2009predictive}.  
Many studies have also been conducted to increase our understanding of such systems, by providing measures for predicting the team's performance \citep{lewis2013human, sycara2012human, zheng2011many, crandall2003towards}.

A different structure of single-human -- multi-robot systems is one where the human user is not in the role of operator or supervisor.  For instance, a team of robots can be used to provide navigation instructions to the human user \citep{yedidsion2019optimal}, cooperatively navigate with them in potentially dangerous situations \citep{penders2011robot, saez2011experiments}, or efficiently guide them in an indoor environment \citep{tan2019one, khandelwal2015leading}.  
Another application of this team size is seen in studies like Emotional Storytelling in the Classroom \citep{leite2015emotional}, where multiple robots team up to execute a storytelling task in front of a student.  Several systems have employed a team of robot actors to have more effective storytelling or drama \citep{swaminathan2021robots, murphy2011midsummer}.

%%%%%%%%%%%%%%%%%%%%%%%%%%%%%%%%%%%%%%%%%%%%%%%%%%%%%%%%%%%%%%%%%
\subsubsection{Multi-human -- single-robot} 
Collaborative systems with multiple human teammates interacting with a single robot have traditionally been used in applications like search and exploration and operating unmanned aerial vehicles (UAVs) where several humans collaborate among themselves to manage a robot \citep{bruemmer2005shared, murphy2004human}.  In such settings, humans take different roles (e.g., pilots, sensor operators etc.) and manage separate components of the robot's operation \citep{murphy2008crew, drury2006decomposition}.  These applications have been important in scenarios when a single human is unable to manage the robot, or when robot failure can result in critical degradation of human's ability to supervise multiple robots \citep{mccarley2005human}.
There are also several systems in which a robot is deployed in an environment with multiple humans it can interact with, either to assist them \citep{claure2020multi,carlson2015team} or to get assistance itself \citep{rosenthal2012someone}.  Another interesting application of such team size is seen in systems where a robot is used as a resource distribution agent within a team of multiple humans \citep{jung2020robot, claure2020multi}, or as a moderator in a group interaction setting \citep{short2017robot, vazquez2016maintaining}.
    
More recently, with increasing research in social robotics, we see a rapid emergence of robotic systems in classroom and public settings.  Starting with \textit{triadic} interactions \citep{salam2016fully, wainer2014using, johansson2013head}, research has presented robots collaborating with multiple users in applications like autism therapy \citep{kim2013social, kozima2005interactive, dautenhahn2003roles}, education \citep{fernandez2020analysing, chandra2016children, tanaka2015pepper}, interactions in public spaces \citep{fortunati2018multiple} and other multiple human-robot social interactions \citep{nanavati2020autonomously, foster2012two}.  
For a more detailed review of systems with robots interacting with a group of humans, readers are encouraged to refer to \citep{sebo2020robots}.

\subsubsection{Multi-human -- multi-robot} 
This category consists of systems having both more than one human and more than one robot in the team.  This team size results in systems that are possibly the most challenging to understand and control due to an exponential growth in uncertainty and the number of states that the system can be in \citep{dahiya2022scalable}.  Such systems have been used in military applications where a human team consisting of members of different roles (pilots, supervisor etc.) needs to coordinate with a team of (semi)autonomous robots to achieve task goals \citep{ramchurn2015study, freedy2008multiagent}. 

Multi-human -- multi-robot team composition is also seen in applications like search-and-rescue tasks \citep{kruijff2014experience, lewis2011process, lee2010teams} and other tasks involving supervisory control of heterogeneous human-robot teams \citep{patel2020improving, driewer2007design, bradshaw2004teamwork}.
% explores the applications where multiple distributed humans interact with robots and with software agents that coordinate mission plans, human activities, and system resources.

Multi-human -- multi-robot teams have also been discussed in several theoretical/computational studies addressing the task allocation and operator scheduling problems \citep{dahiya2022scalable, lippi2021mixed, ijtsma2019computational, malvankar2015optimal}.  These studies discuss different methods for solving robot control, or assisting human decision-making in a multi-agent setting.  As an example, \citet{mina2020adaptive} present a framework for adaptive workload allocation based on agents' health conditions and work performances.  \citet{hari2020approximation} present an approximation algorithm for task scheduling and sequencing, so that humans and robots are able to work on those tasks together when necessary.  
Another application area for such systems is seen more recently in social interaction settings where robots are included as social partners in a group of humans to understand the socio-emotional aspects of such teams \citep{oliveira2020looking, correia2018group, iqbal2017coordination}, and in a classroom setting enabling group learning among students \citep{alves2019empathic, leite2016autonomous, leite2015emotional}.

\subsection{Team Composition}
\begin{figure*}[]
  \centering
  \subfigure[]{\includegraphics[height=4.5cm]{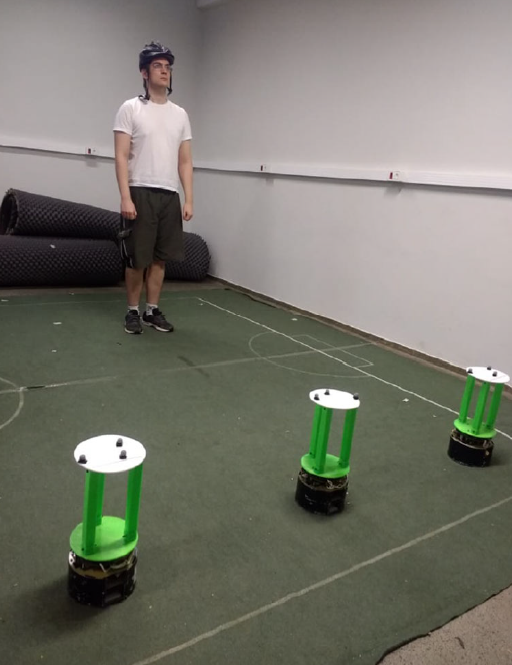}}\;
  \subfigure[]{\includegraphics[height=4cm]{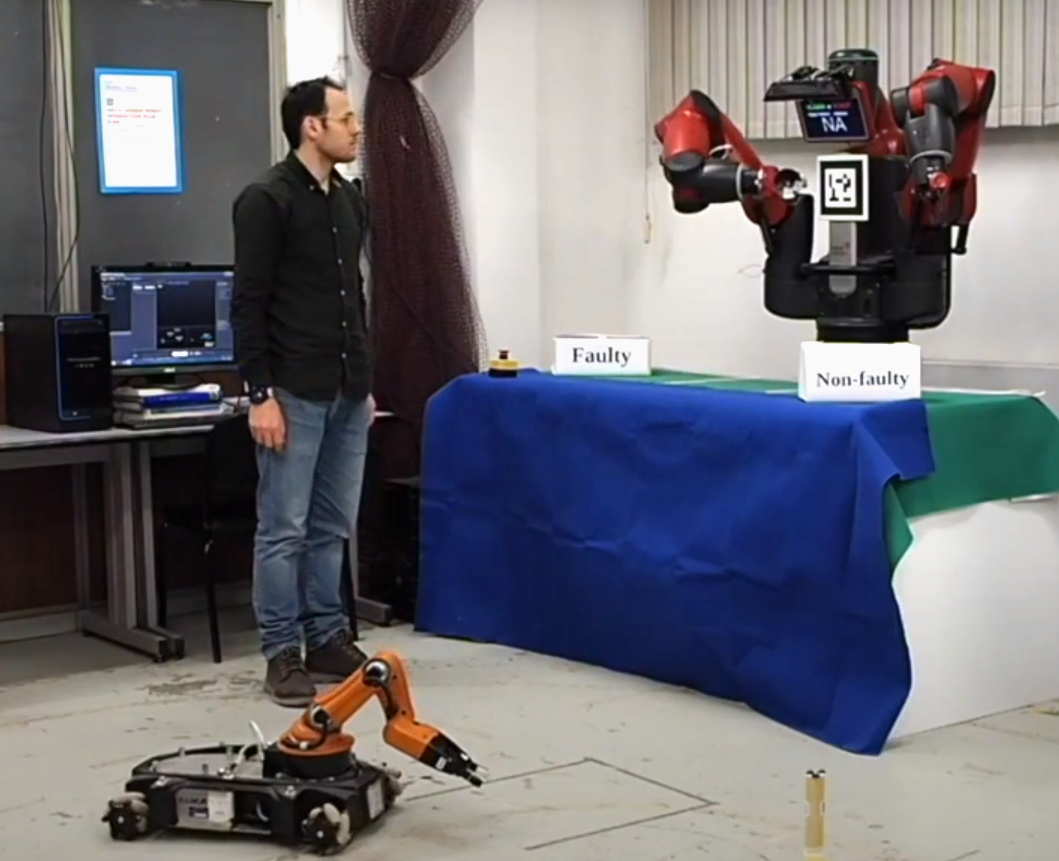}}\;
  \subfigure[]{\includegraphics[height=4cm]{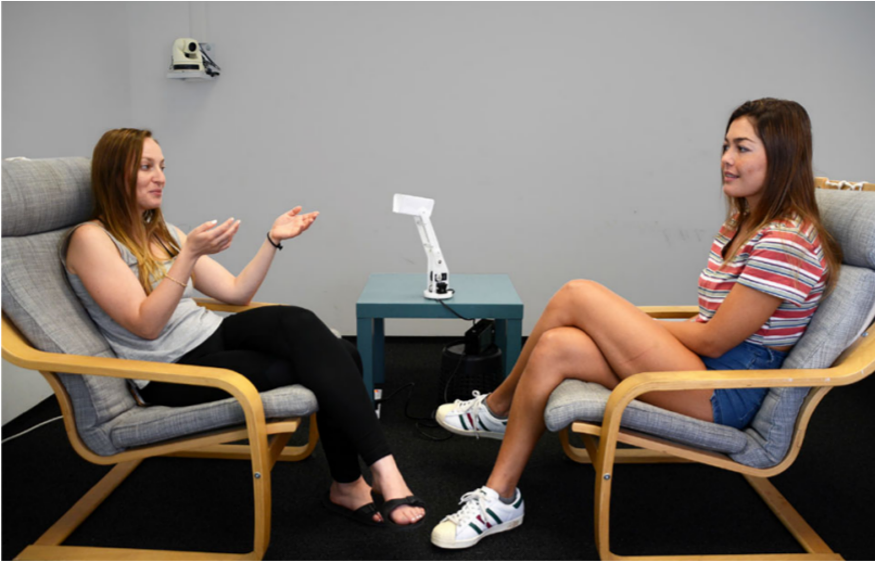}}\;
  \subfigure[]{\includegraphics[height=4.5cm]{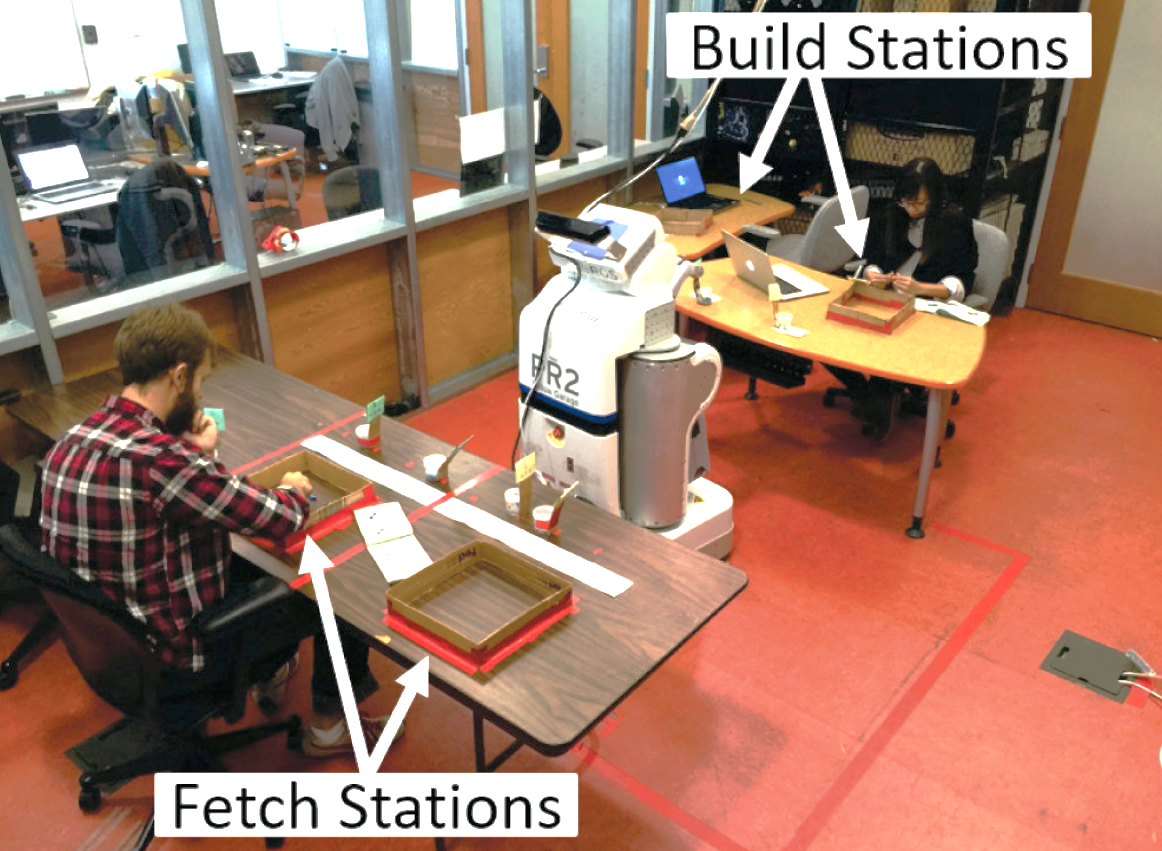}}
  \vspace{-0.75\baselineskip}
  \caption{Human-robot teams with different composition of agents.  a) Multiple homogeneous robots aimed to socially navigate around a human \citep{batista2020socially}; b) A human operator working with two heterogeneous robots in a product inspection task \citep{karami2020task}; c) A robot (in the centre) aiming to provide emotional support during interaction between two humans \citep{erel2021enhancing}; d) Two humans in a role of a leader and an assistant, working with a robot co-leader on a fetching and building task \citep{gombolay2017computational}.
  %d) Children in the role of a teacher and a learner interacting with an educational robot \citep{chandra2016children}.
  }
    \label{fig:team_comp_images}
\end{figure*}

Besides the number of humans and robots, another important characteristic of a human-robot team is the team composition.  This pertains to the aspect of homogeneity (or lack thereof) among the agents, be it humans or robots.  Homogeneity in the case of robots may simply indicate whether there are different types of robots present in the team, with different hardware design \citep{sellner2006coordinated}, manipulation capabilities \citep{karami2020task} or interaction interfaces \citep{kruijff2014experience}.  In the case of humans, \textit{presence} of different roles, capabilities or authority can introduce heterogeneity among team members \citep{chandra2016children, gombolay2015decision}.

Homogeneous team composition is commonly seen in applications where agents are primarily identified as a part of a group rather than individually, such as robots in a swarm \citep{kolling2015human}, or humans in a crowd \citep{chen2019crowd, fortunati2018multiple}.  Applications concerning a group of robots navigating among humans are also mainly making use of homogeneous robots, as shown in Fig.~\ref{fig:team_comp_images}(a) \citep{batista2020socially}.  In the system presented in Fig.~\ref{fig:team_size}(b), the authors investigate the effects of multiple robots on the human participant.  Since the study focuses on the number of robots instead of their individual identities, homogeneous robots are used in the system \citep{podevijn2016investigating}.  In addition to this, homogeneous robots are also useful in applications where each robot is required to work on similar tasks, such as object transfers in warehouse operations \citep{rosenfeld2016human}.
In most social HRI applications, humans are present as equal members of the group and are thus considered homogeneous in their composition, e.g., \citet{foster2012two, erel2021enhancing}.  Heterogeneous robot teams are seen in applications where robots with different manipulators or mobility are required, e.g., \citet{karami2020task, tan2019one}.  In military applications, teams of heterogeneous humans (having different roles, responsibilities and authority) have been used to control one or more complex robots, e.g., \citet{freedy2008multiagent}.

Homogeneity in a human-robot team greatly influences the type, level and efficiency of the interactions \citep{wang2009search}.  For instance, when managing a team of robots, the human operator needs to put in more interaction effort when the team is homogeneous as compared to the one with heterogeneous robots \citep{lewis2013human, goodrich2005task}, and this may lead to an increase in perceived workload and decrease in situational awareness \citep{adams2009multiple, humphrey2007assessing}.  Having homogeneous robots may also lead to a simpler interaction interface \citep{yanco2004classifying}.  However, heterogeneous robot teams may allow to use specific robot capabilities to carry out a variety of operations \citep{suh2009design, sellner2006coordinated}.  It is generally agreed that control and operation of a team of heterogeneous agents is more demanding compared to a  homogeneous team, therefore, the literature offers a variety of studies presenting efficient control strategies for the control of heterogeneous multi-robot teams \citep{saribatur2019finding, rosa2018adaptive, saribatur2014cognitive}.

Likewise, when more than one human is involved in the interactions, differences or similarities among the humans may govern system dynamics \citep{wang2008assessing}.  One common source of introducing heterogeneity among human teammates is through associating different roles with different humans \citep{li2021influencing, kruijff2014experience}.  Earlier studies have presented a taxonomy with different roles (Supervisor, Operator, Peer, etc.) that a robot can assume in an HRI system \citep{scholtz2003theory} and similar roles can also be assigned to human teammates.  There are several systems presenting industrial-oriented tasks that involve a team of heterogeneous humans in roles of supervisors and assistants \citep{gombolay2015decision, murphy2008crew, drury2006decomposition}.  This team composition has also found applications in social/educational robotics, e.g., in systems enabling robots to engage with users from different generations \citep{joshi2019robots, short2017understanding}, or when humans have distinct roles/jobs in the group \citep{taheri2018human}.  

\subsection*{Key Observations}
Considering these team characteristics, it is observed that a substantial amount of work has been done on one-to-many systems (single-human -- multi-robot and multi-human -- single-robot), while systems with both multiple humans and multiple robots are emerging (mainly in the planning and task scheduling literature).  As collaborative robots are envisioned to work together with and among humans, it is essential to further the research that facilitate robots' interactions in systems with a larger number of humans.

Under all types of team sizes, we find systems where humans play different roles in reference to the robots; from supervisors of multiple robots to peers in a social group to information seekers in a public setting.  Robots also show such variety of roles themselves.  
In the past, multiple humans were required to control a single unmanned vehicle, especially in safety-critical settings (e.g., military use).  However, with advancements in robot autonomy, such systems are fading away from the literature and more systems have emerged where a single human is able to supervise a large number of robots.  Human supervision of multiple robots has its own challenges, such as limited human cognitive capabilities, increased workload and decreased situational awareness.  \blu{In later sections, we discuss several studies working towards addressing these issues, including designing of decision support systems, improving robots' motion legibility and predictability, and building intuitive interaction interfaces}.  We expect more such studies to emerge as HRI systems with large number of agents are becoming increasingly popular.
% 
% There is still a lot of work to be done on HRI systems to enable seamless interactions among multiple humans and multiple robots.  
Even though we find many examples of research on systems with heterogeneous robots, implementation of such systems in a social setting or in human-assisting applications is still limited.  

% %%%%%%%%%%%%%%%%%%%%%%%%%%%%%%%%%%%%%%%%%%%%%%%%%%%%%%%%%%%%%%%%%%%%%%%%%%%%%%%%%%
\section{Interaction Style}
\label{sec:interaction_style}
\begin{table*}
     \begin{center}
     \begin{tabular}{ p{0.5cm} | p{3.5cm} | c | p{8cm} |}
% \toprule{2-4}
\cmidrule[0.07em]{2-4}
      & Study & Interaction Graph & Remarks \\ 
\cmidrule(lr){2-4}
         % \hline
         
    \vspace*{\fill}(a)\vspace*{\fill} & \vspace*{\fill} \citet{short2017robot, jung2020robot} \vspace*{\fill}
      &
    \raisebox{5pt-\totalheight}{\includegraphics[width=0.15\textwidth]{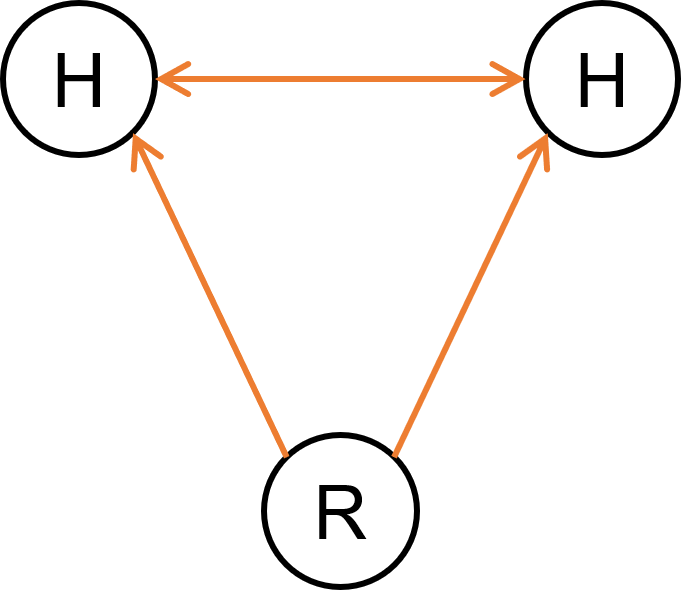}}
      & 
      \vspace*{\fill}
      Robot actions are independent of human teammates.  Humans decide their actions based on robot's actions and potential discussion with the other human teammate.
      \vspace*{\fill}
      \\ 
\cmidrule(lr){2-4}
     \vspace*{\fill}(b) & \vspace*{\fill}\citet{claure2020multi}\vspace*{\fill}
      & 
      \vspace*{\fill}\raisebox{5pt-\totalheight}{\includegraphics[width=0.15\textwidth]{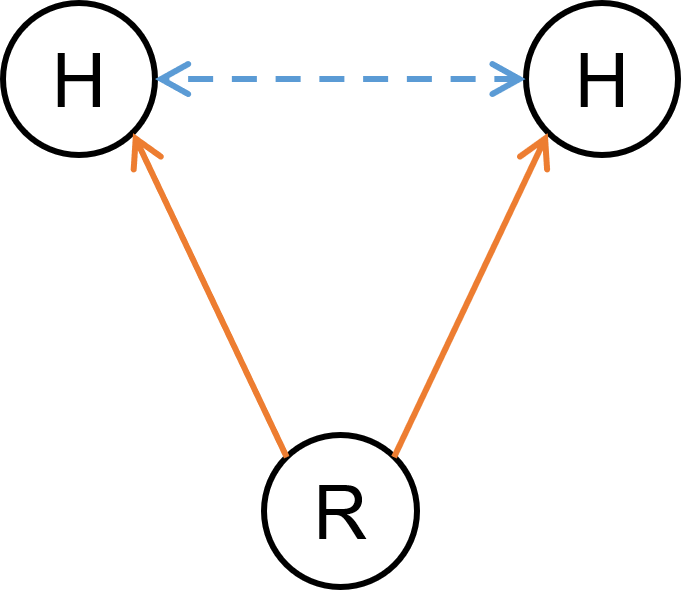}}\vspace*{\fill}
      & \vspace*{\fill}
      Robot actions are independent of human teammates.  Humans decide their actions based on robot's actions, and observing actions of the other human teammate (without an explicit communication). \vspace*{\fill}
      \\ 
\cmidrule(lr){2-4}
     \vspace*{\fill}(c)\vspace*{\fill} & \vspace*{\fill}\citet{tan2019one}\vspace*{\fill}
      & 
      \vspace*{\fill}\raisebox{5pt-\totalheight}{\includegraphics[width=0.15\textwidth]{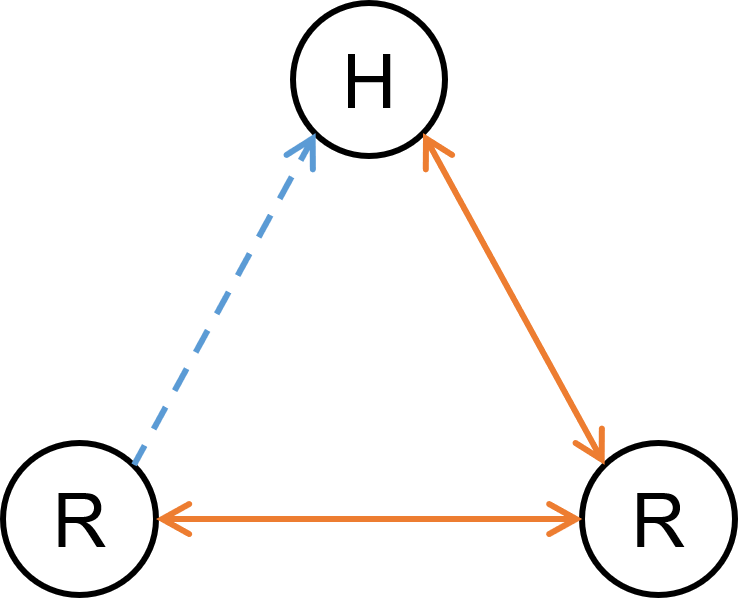}}\vspace*{\fill}
      & \vspace*{\fill}
      A human user directly interacts with one of the two robots, which then conveys the required information to the second robot while the user observes the two robots.\vspace*{\fill}
      \\ 
\cmidrule(lr){2-4}
    \vspace*{\fill}(d)\vspace*{\fill} & \vspace*{\fill}\citet{swamy2020scaled, rosenfeld2017intelligent}\vspace*{\fill}
      &
      \vspace*{\fill}\raisebox{5pt-\totalheight}{\includegraphics[width=0.15\textwidth]{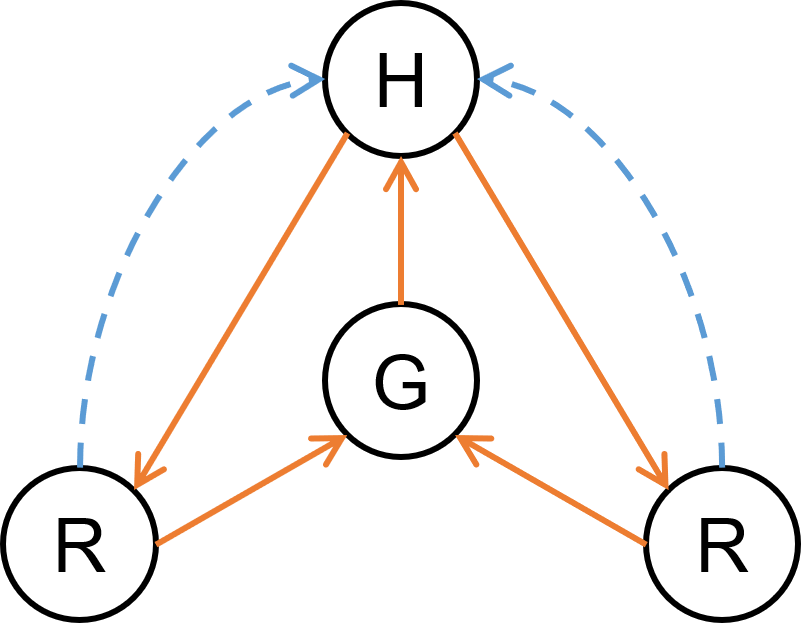}}\vspace*{\fill}
      & 
      \vspace*{\fill}A human operator controls each robot individually.  Human's selection of robot is decided either by observation of robots' states or based on the suggestion shown by the interface (G).\vspace*{\fill}
      \\ 
\cmidrule(lr){2-4}
    \vspace*{\fill}(e)\vspace*{\fill} & \vspace*{\fill}\citet{dahiya2022scalable} \vspace*{\fill}
      &
     \vspace*{\fill}\raisebox{5pt-\totalheight}{\includegraphics[width=0.15\textwidth]{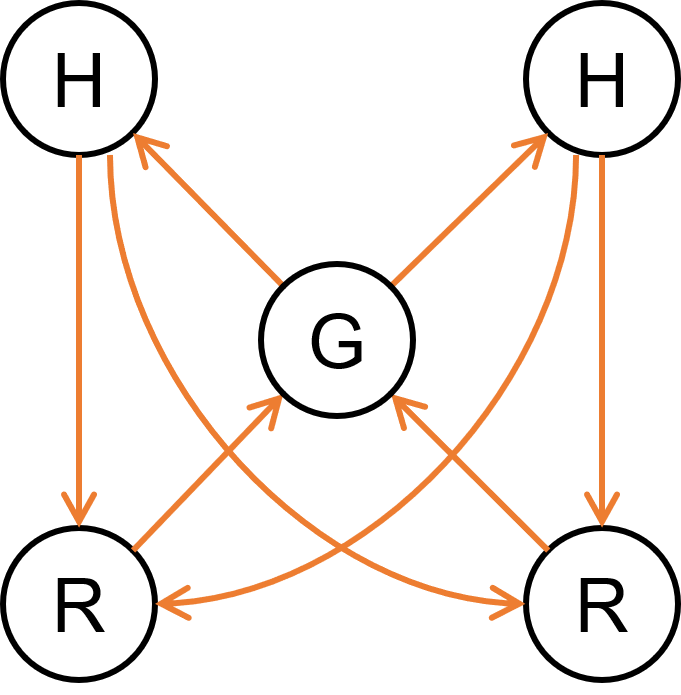}}\vspace*{\fill}
      & 
      \vspace*{\fill}A central computational entity (G) collects data of all robots and then allocates each human operator to the robots that need assistance.  Operators directly assist each robot individually.  Operators are allocated only by the central computational entity and not by any direct interaction from the robots.\vspace*{\fill}
      \\ 
\cmidrule(lr){2-4}
    \vspace*{\fill}(f)\vspace*{\fill} & \vspace*{\fill}\citet{patel2020improving}\vspace*{\fill}
    &
     \vspace*{\fill}\raisebox{5pt-\totalheight}{\includegraphics[width=0.15\textwidth]{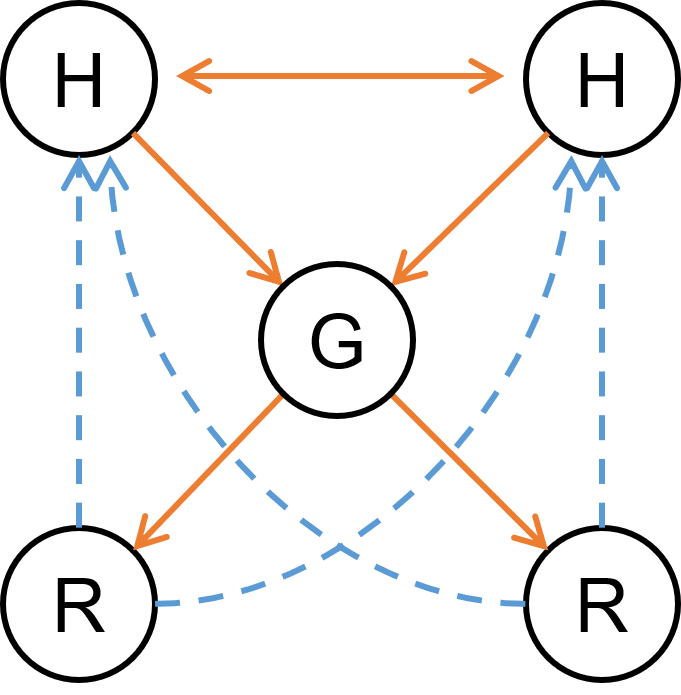}}\vspace*{\fill}
      & 
      \vspace*{\fill}Commands of multiple human operators are converted to actions required by individual robots.  Operators observe each robot directly and can also interact with each other to resolve conflicts. \vspace*{\fill}
      \\ 
\cmidrule(lr){2-4}
    \vspace*{\fill}(g)\vspace*{\fill} & \vspace*{\fill}\citet{correia2018group}\vspace*{\fill}
    &
      \vspace*{\fill}\raisebox{5pt-\totalheight}{\includegraphics[width=0.15\textwidth]{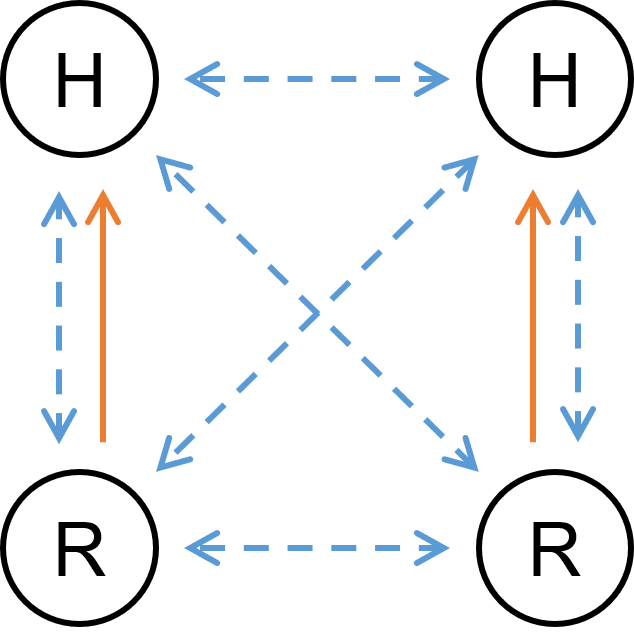}}\vspace*{\fill}
      &
      \vspace*{\fill}Two teams, each consisting of one human and one robot, play a competitive card game.  Each agent decides on their actions based on observations of other agents' actions.  The only direct communication in the system occurs when a robot speaks to its partner to convey its emotions.\vspace*{\fill}
      \\ 
% \bottomrule
\cmidrule[0.07em]{2-4}
      \end{tabular}
      \caption{Examples of multi-agent human-robot systems with respective interaction graphs.  The orange solid arrows signify direct interactions between agents (e.g., explicit communication), while dashed blue arrows denote the indirect interactions (observational).  
      A maximum of two humans and robots are shown in the figures but interaction graphs can be drawn for larger number of agents in a similar way.
      \blu{Note that this table only shows examples of possible interaction graphs and is not an exhaustive list.}}
      \label{tbl:comm_models}
      \end{center}
\end{table*}

\begin{figure*}[htp]
  \centering
  \subfigure{\includegraphics[height=4cm]{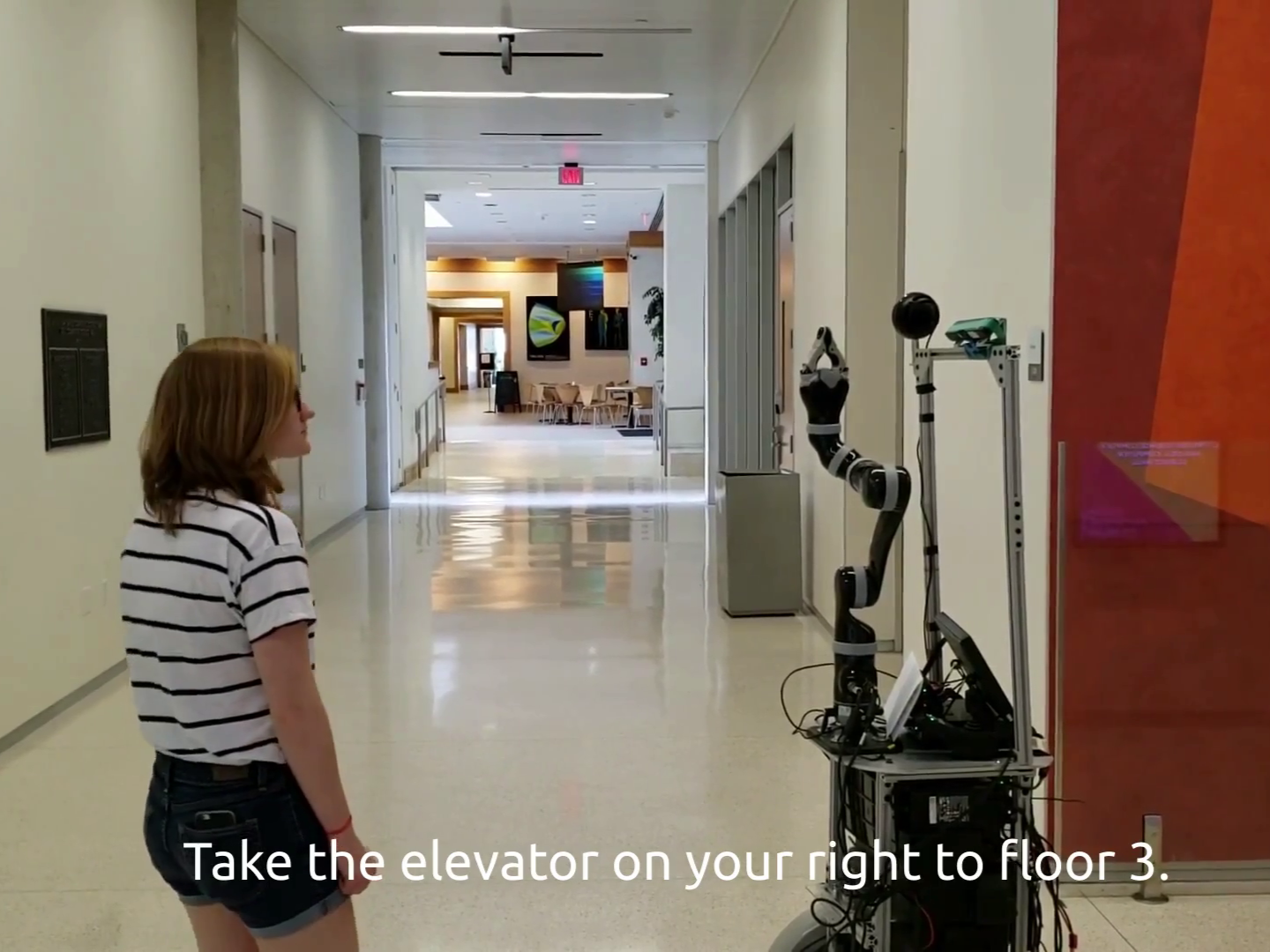}}\;
  \subfigure{\includegraphics[height=4cm]{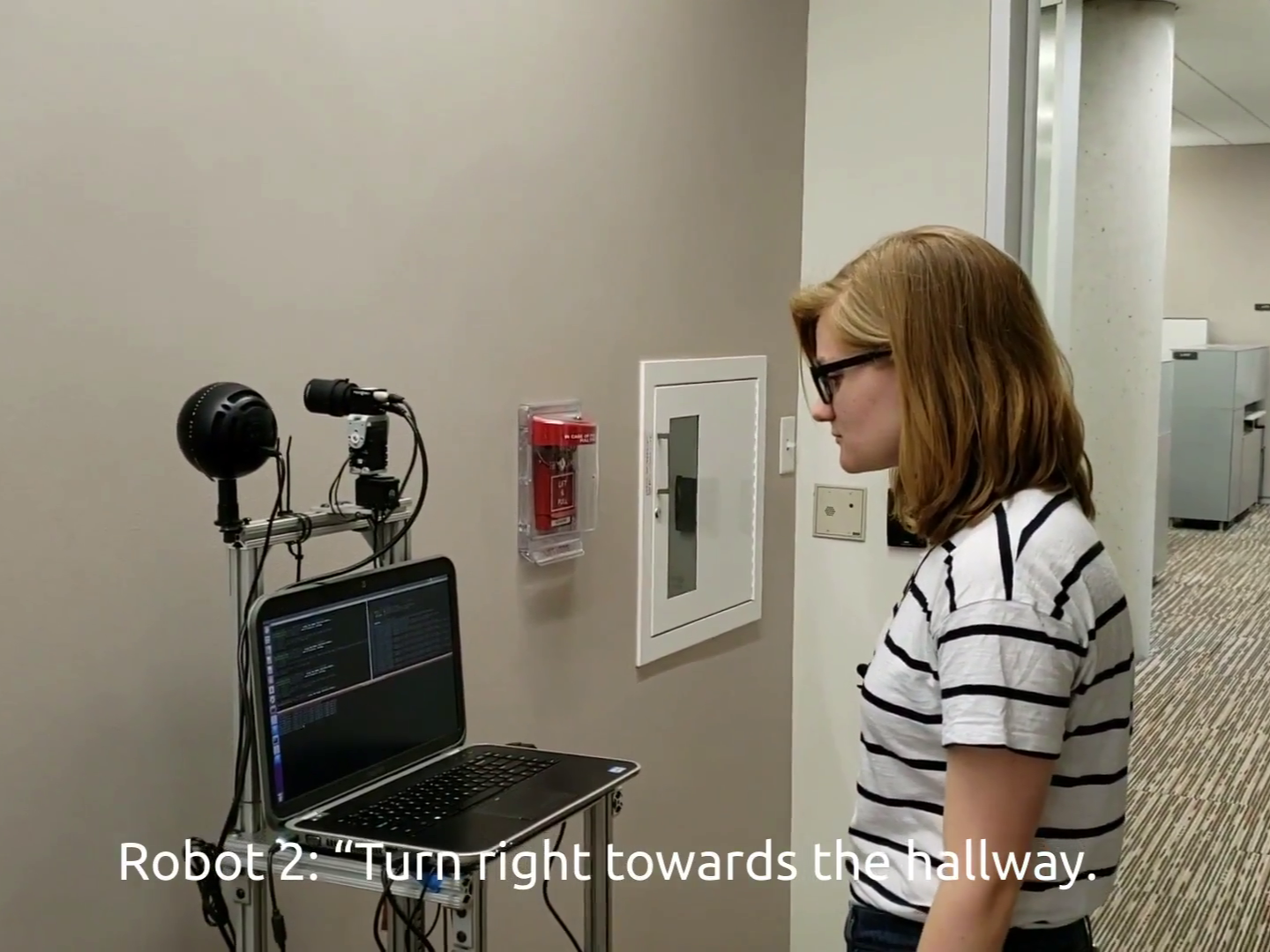}}\;
  \subfigure{\includegraphics[height=4cm]{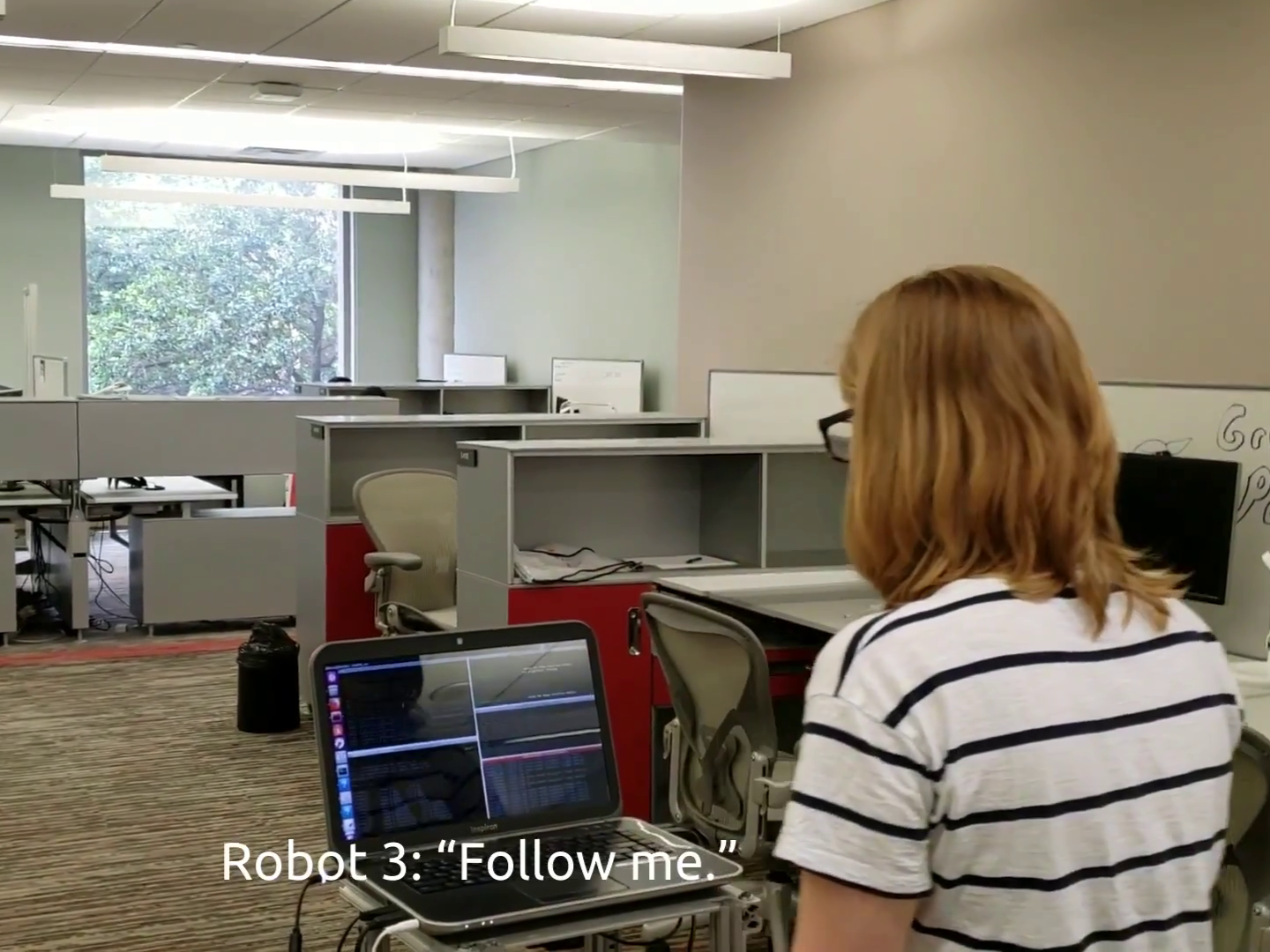}}\;
%   \vspace{-0.5\baselineskip}
  \caption{Example of a one-to-one interaction model in a single-human -- multi-robot system \citep{yedidsion2019optimal}.  Multiple robots connected via a network to guide a human user who interacts with the robots located in different locations.}
    \label{fig:121_interaction_yedidsion}
\end{figure*}
In the context of human-robot systems, interaction style is a broad term that can be used to refer to different aspects of interaction such as the modes of communication among agents \citep{feine2019taxonomy}, interaction models \citep{fortunati2018multiple} and the interaction interface \citep{gromov2016wearable, rule2012designing}.
Interaction style also includes the method of communication (verbal/non-verbal) \citep{mavridis2015review, stiefelhagen2004natural}, expression of affect \citep{schermerhorn2011disentangling} and spatial relationship \citep{chen2007human, huttenrauch2006investigating}.   

Interactions in multi-agent human-robot systems differ from those in dyadic systems in three ways: \\
1) First, with multiple agents present, it is possible to have interactions within a group of agents, viz.  Human-Human Interactions, e.g., \citet{kruijff2014experience} and Robot-Robot Interactions, e.g., \citet{williams2015covert}. \\
2) Second, in addition to the one-to-one interactions seen between humans and robots in dyadic HRI systems, one-to-many interactions are also realizable in multi-agent HRI systems, e.g.,  \citet{fortunati2018multiple}.    \\
3) Third, in multi-agent systems, there are additional types of interactions possible among the agents.  
\citet{patel2021direct} discussed the differences between `direct' and `indirect' interactions between two humans in a system in which they can communicate either via verbal communication or through the interface.  \citet{che2020efficient} investigated the role of `explicit' and `implicit' communication in social navigation.
\citet{abrams2020c} used the terms `the group' and `the observer' to distinguish the two perspectives of measuring group cohesion in a multi-agent HRI setting. 

Such distinctions of interaction types are useful to better understand HRI systems and can be applied to improve the interaction outcome.  In the context of multi-agent HRI systems, we find it useful to distinguish `direct' and `indirect' interactions. 
Direct interaction between a sender and recipient occurs when the sender actively (i.e., intentionally and explicitly) communicates to the recipient using any mode, verbal or non-verbal.
A third party may also receive information from this direct communication, and we refer to this ``eavesdropping" as indirect interaction.
Taking the example of a study presented in \citet{tan2019one}, a robot trying to transfer task information to another robot (e.g., through speech) is an example of direct interaction, whereas a human observing the robots interacting with each other is an example of indirect interaction.

% Unlike the dyadic setting, an agent in multi-agent systems cannot only observe other agents but also observe their interactions, e.g., a human being influenced by the interaction between two robots in the system \citep{tan2019one}.  
% \blu{Note that, we do not distinguish between intentional and unintentional observations, rather we differentiate based on whether the observations are intended for the observer or not.  For example, a robot trying to do legible movements for a human teammate to observe is an example of active interaction, whereas a human observing the robots and/or humans interacting with each other is an example of passive interaction.}  
Making this distinction can help us understand if and how much the agents in the system are actively trying to communicate with each other, or if they are primarily just co-existing in the same environment while observing each other.  
% As agents observe interactions between other agents, there emerges another kind of interaction called \textit{indirect} interactions.  
Modelling the indirect interactions is necessary to understand how direct interactions between any two agents can affect the behavior of others, e.g., \citet{rifinski2021human, tan2019one, short2017robot}.  

Examples of systems with different types of interactions are given in Table~\ref{tbl:comm_models}.  
In the remainder of this section, we discuss the above-mentioned differences under two key aspects of interaction style: 1) Interaction model present, and 2) Communication modalities used in the system.

\subsection{Interaction Models}
The presence of multiple agents doesn't necessarily mean that each interaction in the system also involves multiple agents.
Interactions in a multi-agent HRI system can be implemented using different-sized communication channels (i.e., number of agents connected via a single interaction).  
In a multi-agent system, an agent has the capability of interacting with one or multiple agents at the same time, resulting in one-to-one or one-to-many interaction models \citep{fortunati2018multiple}.  Examples of several multi-agent HRI systems with different interaction models are shown in Table~\ref{tbl:comm_models} using the interaction graph structure defined in Section~\ref{sec:general_classification}.  These interaction models are an extension of the interaction types presented in \citet{yanco2004classifying}.  The presented models are able to include a variety of interaction models that are possible in multi-agent HRI systems, especially in social interactions.  In addition, these models let us specify the direction of information flow between agents in the system and can illustrate differences between human-to-robot and robot-to-human interactions.

\textbf{Note:} These models, in their basic form (Fig.~\ref{fig:interaction_arrows}), do not describe many aspects of the complete system (e.g., they don't show embodiment, roles or heterogeneity among agents and modes of communication etc.).  However, it is possible to add some of this information with slight modifications.  In Table~\ref{tbl:comm_models}, we show one such additional information, i.e., whether an interaction is direct or indirect using solid and dashed arrows respectively. 

As observed from Table~\ref{tbl:comm_models}, interactions between different pairs of agents in a system can be implemented using different models (one-to-one, one-to-many etc.).  It is also possible that the interaction model varies with the type of agents involved, i.e. different models for human-to-robot, robot-to-human, robot-to-robot and human-to-human interactions.  For instance, a human may be able to give commands to a single robot at a time (one-to-one) but may observe behavior and receive messages/information from all robots at once (one-to-many). \citep{swamy2020scaled, khasawneh2019human}.
These models are discussed below under three categories denoting how many agents are connected at each end of the communication channel.

\subsubsection{One-to-one}

The one-to-one interaction model is similar to the one found in \textit{dyadic} human-robot systems, where one robot is interacting with one human teammate at any given time, even though there may be other agents present in the environment.  This form of interaction is commonly seen in systems where one can afford the commands/instructions for an agent to be independent from others (e.g., in systems where coordination among agents is not required).  A common application is seen in systems where a human operator manages a team of multiple robots by giving each of them separate commands.  
% In such systems, it becomes a priority to facilitate the human by maintaining their situational awareness or by helping them switch their attention across robots.
% Several studies have used the concept of sliding autonomy to enable such a human input in the multi-robot team \citep{dias2008sliding, lewis2013human}.  
While a fleet of autonomous robots acts in separate environments, a human operator can monitor their states remotely, and can intervene to help a robot via teleoperation when the robot encounters a challenging state \citep{swamy2020scaled, rosenfeld2017intelligent}.

Human commands to a robot swarm can be implemented both with a one-to-one and a one-to-many interaction model.  One common strategy under the one-to-one model is the leader-follower approach where human users directly control the leader robots, and the control signals for rest of the swarm are governed by the leaders’ motions \citep{setter2015haptic}.  Human-swarm interaction can also be designed to follow a one-to-many interaction model as discussed later in Sec.~\ref{sec:12M_comm}.

% In case of direct interactions (ones resulting directly from actions and communication of an agent), 
One can easily identify the interaction model by checking if an interactive action of an agent is targeted at one or multiple agents.  For instance, in the system shown in Fig.~\ref{fig:interaction_arrows}, even though the robot interacts with multiple humans throughout the duration of its task, at any given point the robot's actions are directed towards a single human \citep{jung2020robot}.  Figure~\ref{fig:team_comp_images}(b) shows a system consisting of two robots with a human co-worker.  In this system, the human interacts with each robot separately and none of the interactions involve all  three agents at once \citep{karami2020task}. 
We also see examples of such HRI systems in an office setting, where even though multiple humans or robots are present, each interaction is between a single human and a single robot. An example can be a system with a robot going around in an office asking help from human users at different times \citep{rosenthal2012someone}.  In Fig.~\ref{fig:121_interaction_yedidsion}, we see a person being guided from one robot to another by an indoor guidance system, where the human interacts with only a single robot at a time \citep{yedidsion2019optimal}.

In settings where agents are located in the same environment, the system also has indirect interactions, occurring when an agent observes other agents and interactions are happening between those agents.  
% When humans are located with other agents (humans or robots) in the same environment, their behavior is influenced by everything else happening around them, be it interactions happening between other agents or non-verbal cues they show.  
These indirect interactions are still following a one-to-one model as information flow from each agent is directed towards the observer.  

\begin{figure*}[htp]
  \centering
  \subfigure[]{\includegraphics[height=4.5cm]{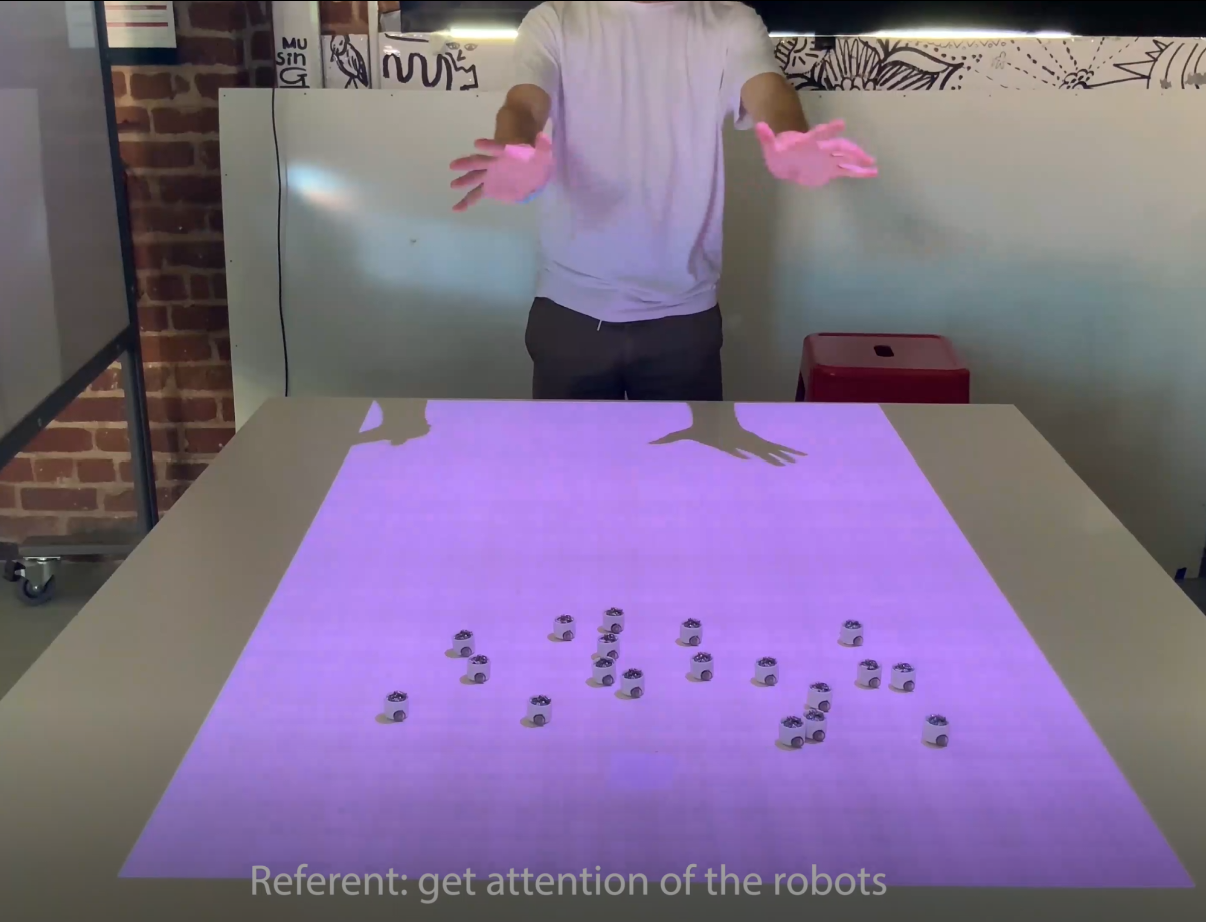}}\;
  \subfigure[]{\includegraphics[height=4.5cm]{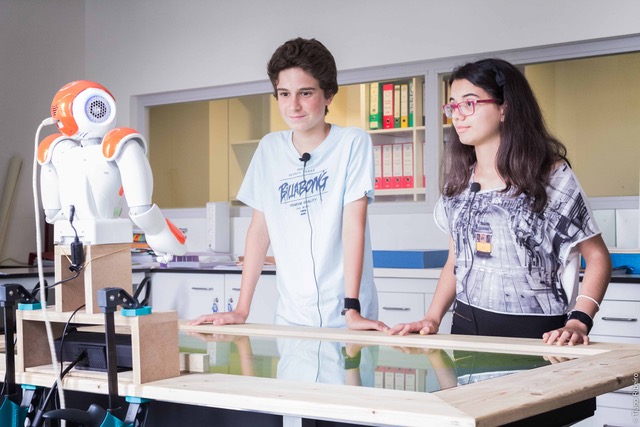}}\;
  \subfigure[]{\includegraphics[height=4.5cm]{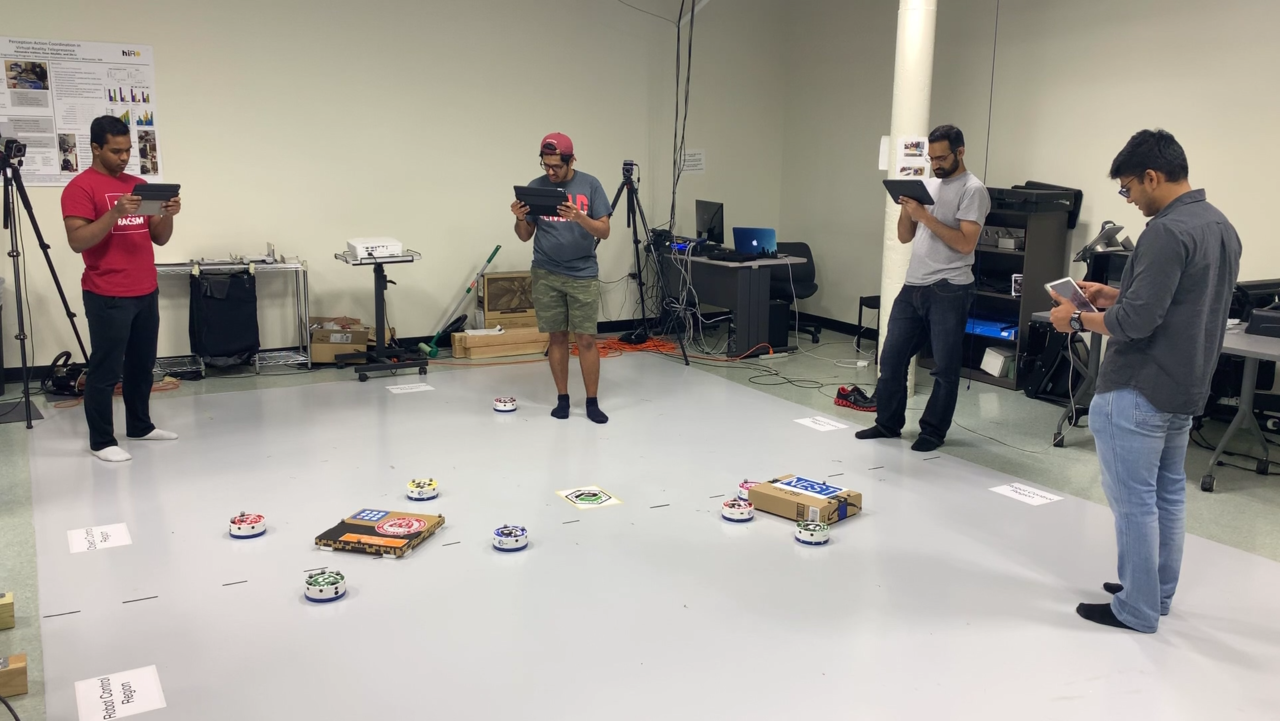}}\;
  \vspace{-0.75\baselineskip}
  \caption{Examples of different interaction models: a) One-to-many: A system where a human controls multiple robots at once using hand gestures \citep{kim2020user}.  \blu{The gestures are interpreted and translated into group commands for the robots.}  b) One-to-many: A robot playing a table-top game where it interacts with two human teammates \citep{alves2019empathic} c) Many-to-many: Multiple humans giving commands to multiple robots using a tablet interface \citep{patel2020improving}, \blu{where each human operator can give commands to multiple robots simultaneously. Any form of conflict in commands is then resolved through the interface or verbal communication among the operators.}}
    \label{fig:interaction_models_images}
\end{figure*}

\subsubsection{One-to-many}
\label{sec:12M_comm}

% \todo{Define one-to-many interactions? What these include and what they don't.}
The one-to-many interaction model can be seen in situations where one robot is directly interacting with multiple humans at once \citep{nanavati2020autonomously, correia2018group, alves2019empathic},  and where one human is simultaneously interacting with multiple robots \citep{lim2018human, guo2009touch, arkin1994integration, jones2001supervisory}.  Note that while some works distinguish between many-to-one and one-to-many models based on the direction of control commands (e.g., \citep{lim2018human}), in this article all such systems are discussed together under a one-to-many model.

A simple way of implementing this interaction model is to use verbal communication or gestures to issue group commands to multiple agents at once, e.g., \citet{kim2020user}.  Consequently, interactions under this model are common in social settings where it naturally finds its utility as the robots and humans are co-present in the same environment.  Studies such as \citep{nanavati2020autonomously, correia2018group} present systems where robots' actions and behavioral cues are directed towards the whole group of human teammates.  Another similar application of this model is seen in classroom settings where a robot acts as a teacher/tutor for a group of students to promote group-based learning \citep{leite2015emotional, alves2019empathic, belpaeme2018social}.  A robot interacting with humans in public space also naturally calls for a one-to-many model \citep{fortunati2018multiple}.

In industrial or military-oriented applications, where humans and robots might be located in separate environments, interactions with one-to-many model are usually implemented using a group command node (see interaction graphs in Table~\ref{tbl:comm_models}; shown as node (G)).  This group command node is an interpreter -- an interface or an algorithm -- that converts communication from one or more agents to information required for each individual agent before relaying that information to its recipient(s).
For example, when controlling multiple ground or aerial robots, a human operator can simply give group commands to the fleet instead of telling each robot what to do, while the interpreter converts this group command to required actions for each robot \citep{lim2018human, ayanian2014controlling}. 
As mentioned earlier, implementation of human interactions with a swarm can also make use of one-to-many model.  In order to make human control efficient, a group command node is usually implemented to convert human intentions/commands into control signals for all robots \citep{kolling2015human, podevijn2013gesturing}.  

\subsubsection{Many-to-many} 
In multi-human -- multi-robot systems, it is also possible to have multiple one-to-many interactions taking place among different agents, resulting in a \textit{many-to-many} interaction model.  This is the most unconstrained interaction model that can be implemented in a human-robot system.  Although it is not very common to find such interaction models in the literature, studies by \citet{patel2021direct, zhang2016optimal, tews2003scalable} have presented interaction interfaces to facilitate many-to-many interactions.  
 
A common way of enabling interactions among the agents in such systems is through a proxy architecture aimed at facilitating collaboration of humans and robots with varying levels of autonomy \citep{mostafa2019adjustable, ramchurn2015study, freedy2008multiagent}.  Under this setting, each agent interacts with a proxy, which can be part of a centralized proxy architecture or connected to a network of proxies.  Such proxies are usually responsible for receiving inputs from different agents, interpreting the messages/commands and then relaying the relevant information to other agents.  Augmented Reality (AR) based interfaces have also been used for making such interactions more user-friendly by providing easier methods of giving commands.  For example, the system shown in Fig.~\ref{fig:interaction_models_images}(c) uses an AR interface on separate tablets to enable multiple users interact with multiple robots \citep{patel2020improving}.
Without such proxies, the human user in the system must resolve any conflicts (in commands/decisions) by themselves and work out a common strategy \citep{hwang2014case}.

Looking at the interaction graphs in Table~\ref{tbl:comm_models}, many-to-many interactions can be represented as a group node (G) having multiple incoming and multiple outgoing edges.  For example, in Table~\ref{tbl:comm_models}(e), a central computational entity -- shown as group node (G) -- collects data from multiple robots and relays allocation information to multiple human operators.  In Table~\ref{tbl:comm_models}(f), the interface enables multiple users  to input control commands and then decides actions for individual robots.
A many-to-many model provides a natural and efficient interaction setup in a multi-agent system as there are minimal constraints on when agents are allowed to communicate with each other, and information from multiple agents can be relayed to respective recipients simultaneously.

%%%%%%%%%%%%%%%%%%%%%%%%%%%%%%%%%%%%%%%%%%%%%%%%%%%%%%%%%%%%%%%%%%%%%%%%%%%%%%%%%%%%%%
\subsection{Communication Modalities and Interfaces}
Communication modalities refer to the modes through which different agents interact in the system (e.g., speech, haptics, screen etc.).  Similar to the conventional dyadic HRI systems, a multi-agent system can have agents communicate through verbal or non-verbal modes, or a combination of multiple modes at once.  The chosen modes of communication depend on the environment, application of the system, proximity of different agents and their interaction capabilities.  Each modality has its own advantages, implementation requirements and limitations.  
As we note below, proximity has a major influence on which communication modalities are feasible in a system to enable efficient communication among humans and robots.  Therefore, to discuss the different types of communication modalities used in multi-agent systems, we find it useful to group the communication modalities based on proximity of the agents.

\begin{figure*}[htp]
  \centering
  \subfigure[]{\includegraphics[height=3.5cm]{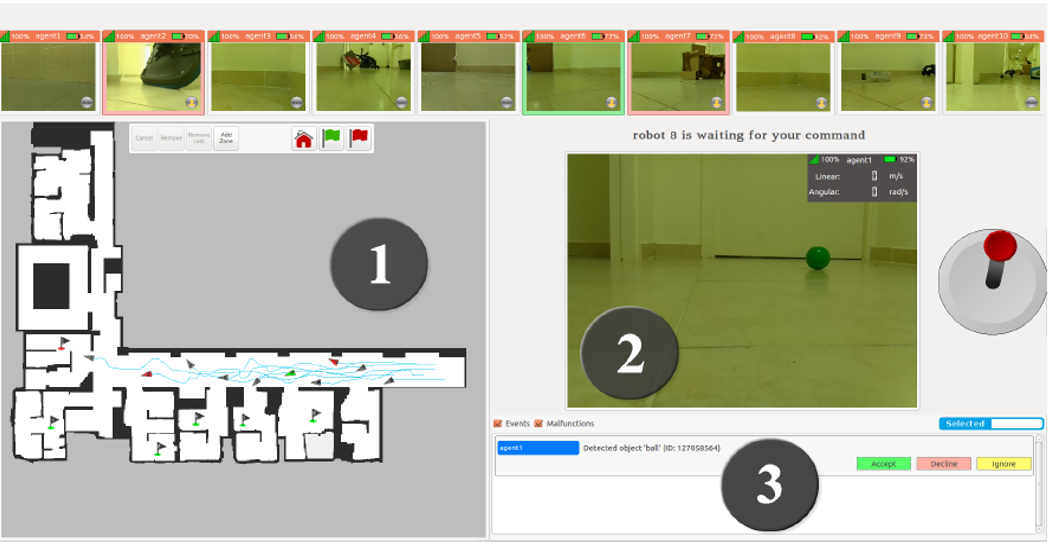}}\;
  \subfigure[]{\includegraphics[height=3.5cm]{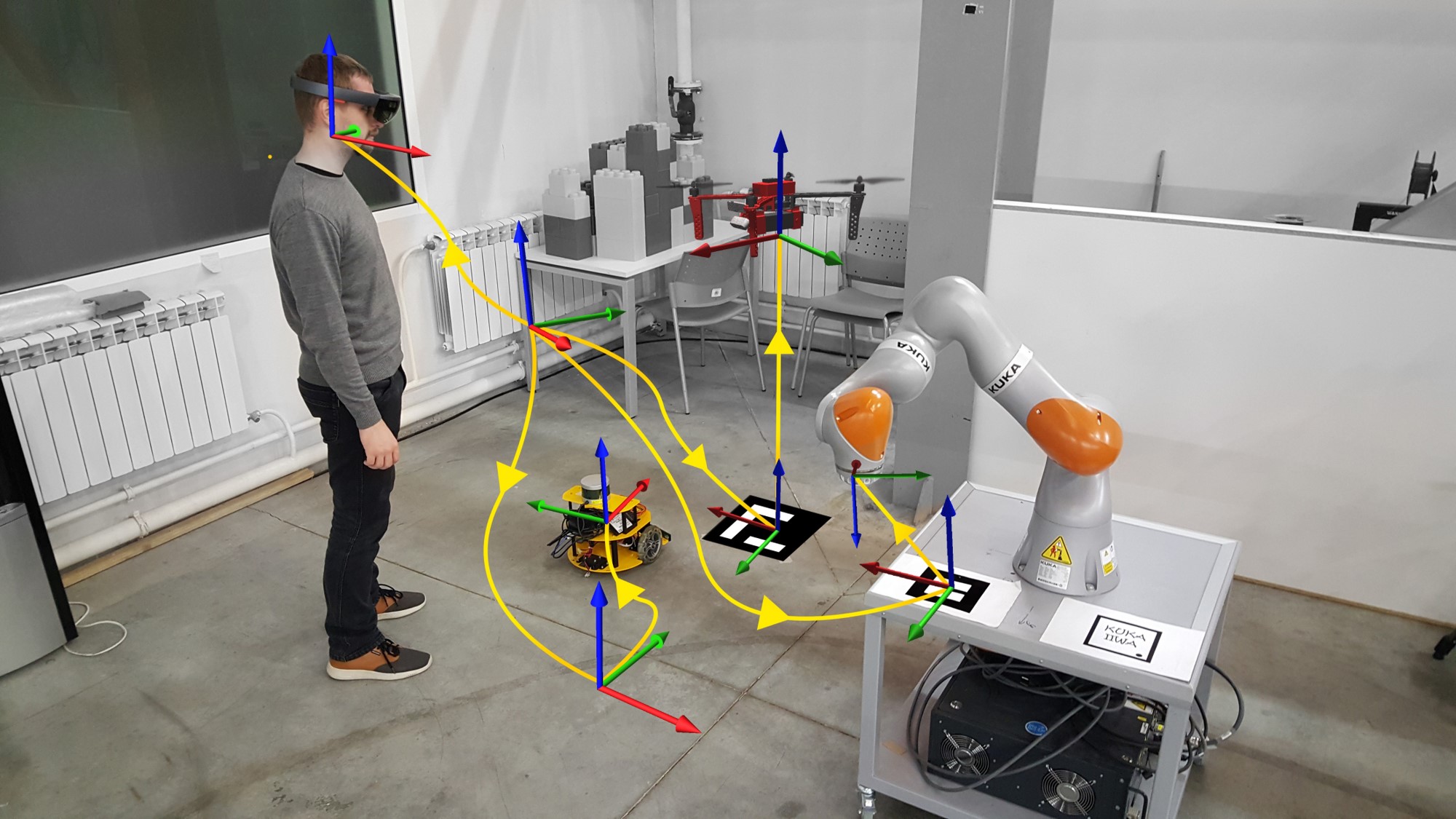}}\;
  \subfigure[]{\includegraphics[height=3.5cm]{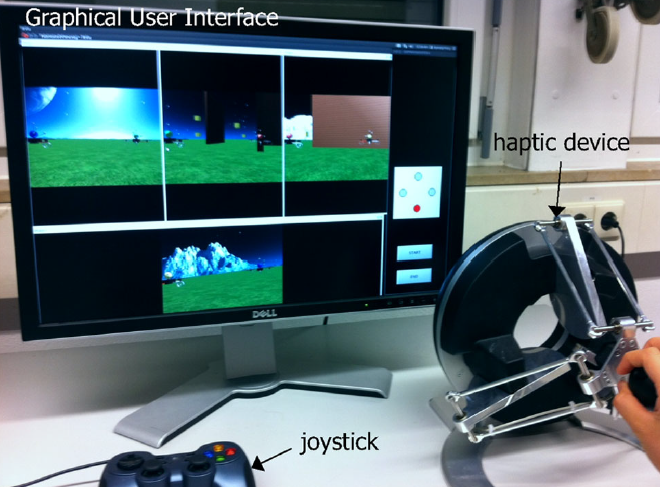}}\;
  \vspace{-0.75\baselineskip}
  \caption{Examples of interface design for control of multiple robots:  a)  Interface of a search-and-rescue task with multiple robots \citep{rosenfeld2017intelligent}. Such screen-based interface designs are common in applications where a human user controls multiple remote robots. b) A mixed-reality based multi-robot control framework. The figure shows relation of coordinate frames of main entities of the system, including one human and three robots \citep{ostanin2021multi}. c) Teleoperation of multiple mobile robots using a joystick and a haptic device \citep{hong2017multimodal}.}
    \label{fig:interfaces}
\end{figure*}

\subsubsection{Remote communication}
For the systems designed to operate in environments that can be potentially dangerous or inaccessible to humans, a screen-based interface is ubiquitous for enabling remote communication among humans and robots.  Human control of a fleet of UAVs \citep{nam2017predicting, drury2006decomposition}, supervision during search and rescue tasks \citep{khasawneh2019human, scholtz2004evaluation, murphy2004human} and teleoperation of underwater robots \citep{wang2014human} are some of the applications that make use of remote communication techniques.

Remotely interacting with robots is shown to result in higher cognitive workload in human users \citep{gittens2021remote} that increases further with an increasing number of robots \citep{fincannon2013best, adams2009multiple}.  Furthermore, communicating plans to team supervisors and maintaining their situational awareness are important challenges faced while collaborating with remote robots \citep{hastie2019challenges}.
Therefore, the development of interaction interfaces that enable efficient and reliable human interaction with multiple remote robots is an important area of research in remote HRI \citep{roldan2017multi, rule2012designing, driewer2007design}.  In systems with a disproportionately high number of robots compared to human operators/supervisors, it becomes difficult for the human users to interact with all robots efficiently due to challenges related to perception, workload and situational awareness \citep{fincannon2013best}.  Therefore, intelligent interface designs are required that can mitigate these challenges \citep{hussein2018towards}.  So far, the literature mostly consists of screen-based interface designs as they are most convenient to implement in such systems as seen in many of the studies discussed in \ref{sec:1hmR}.  Conventionally, many of the multi-robot interfaces have several common design features such as camera feeds of multiple robots in small cards along the screen edge, an enlarged view of one selected robot and a map, or an overview, of all robots in the environment \citep{rosenfeld2017intelligent, chien2018attention} as shown in Fig.~\ref{fig:interfaces}(a).  Recently, immersive interfaces using Virtual Reality (VR) or Augmented Reality (AR) have gained popularity in multi-robot systems \citep{ostanin2021multi, roldan2017multi}, and have been shown to improve operator's situational awareness and decreases workload \citep{frank2017toward}.
Some systems also augment the interaction interface with multiple communication modes (e.g., haptics) to improve interaction outcome \citep{hong2017multimodal} (Fig.~\ref{fig:interfaces}(c)).

\subsubsection{Proximate communication}
In systems where humans and robots are located in the same environment, many more options are available to system designers in terms of choosing modes of communications.  These include communication via auditory channels (speech and non-speech audio), visual channels (gestures, facial expressions, body postures and gaze), and physical channels (touch and force).  
The human guidance system presented by \citet{yedidsion2019optimal} (shown in Fig.~\ref{fig:121_interaction_yedidsion}) uses simple verbal instructions to communicate with the human.  Figure~\ref{fig:interaction_models_images} shows a human user controlling a robot swarm using gestures \citep{kim2020user}.

It is also possible to use multiple modes at once, resulting in a multi-modal communication architecture.
\citet{pourmehr2014you} present a system that uses a combination of haptic and verbal inputs from a human user to control multiple UAVs.  In the system presented by \citet{gromov2016wearable}, a human user can communicate with robots using speech and gestures, while robots provide visual and verbal feedback to the user. 

Proximate communications are more commonly seen (and naturally present) in social robotics as it facilitates more personal interactions required in a social setup.  This is expected as the meaning conveyed through an interaction significantly depends on social cues (verbal and non-verbal), and not solely on the concrete message being communicated \citep{feine2019taxonomy}.  
When humans and robots are co-located in the same environment, the role of non-verbal communications becomes important to consider.   For example, the robot shown in Fig.~\ref{fig:team_comp_images}(c) is used to enhance Human-Human Interaction using only gestures, without interfering with the interaction \citep{erel2021enhancing}.
There have been several other studies that investigate different cues (expressions and movements) as a mode of communicating a robot's intent, emotions and information more clearly in a group \citep{correia2018group, faria2017me}.  Moreover, the spatial placement of agents, in relation to each other and the environment, also affect human behavior in a group setting \citep{rios2015proxemics, yamaoka2009model}.

When the agents are located in the same environment, it is also likely that interaction between any two agents has an effect on behavior and future interactions of other agents.  This is an example of indirect/implicit communication between agents.  Thus,  Human-Human Interactions can be affected by Human-Robot Interactions \citep{joshi2019robots, short2017understanding, kim2013social}.  Likewise, Robot-Robot Interactions can also impact future human interactions with the robots \citep{yang2012effect, tan2019one}, or can affect a human's psychological state \citep{erel2021excluded}.  Moreover, the embodiment or mere presence of a robot may influence human behavior and their perception of the robots \citep{druckman2021best, shiomi2020survey}.

Communication modalities can also vary depending on the agents involved and the direction of communication.  For instance, \citet{berg2019human} present an interaction system where the human to robot communication channel is realized through gestures and eye tracking while the information from the robot to the human is communicated using a projection.  In the system presented by \citet{rosenthal2012someone}, the robot communicates its queries using speech and receives human input using a visual interface on a laptop.  Also, it is common to see different communication modalities between Human-Robot and Human-Human Interactions \citep{kruijff2014experience}.

\subsection{Key Observations}
% \todo{Highlight how interaction among multiple agents is different from one-to-one interactions.}
In the surveyed literature, we looked into different forms of interaction styles present in multi-agent human-robot systems.  In terms of interaction models, a couple of \textit{trends} are noticed.  Even with multiple agents present in the system, the one-to-one interaction model is the most prevalent, signifying that agents mostly interact with only one other agent at a given time.  The one-to-many interaction model has also gained popularity and has been useful to enable more efficient interactions in multi-agent systems.  Such systems are mostly seen in the area of social robotics and co-located settings.  In order to realize more {natural} and \textit{smooth} interactions among humans and robots, a many-to-many model is desirable.  We are beginning to see such implementations and hopefully this will mature into a strong interaction model in human-robot systems.

In regards to communication modalities, the systems where humans are interacting with multiple robots, using a screen-based interaction interface is the most common (and possibly the most practical) one.  With the advent of Virtual Reality (VR) and Augmented Reality (AR) technologies, the future multi-agent Human-Robot Interaction systems may shift from conventional two-dimensional screen-based interfaces to more immersive ones.  It is still unclear whether using VR and AR will improve system usability in diverse settings and more research is required to investigate the applicability of these technologies in multi-agent HRI systems for different applications.  In co-located and social settings, multi-modal communication is becoming more popular and can enable more natural and safer interactions between humans and robots, without the need of enforcing physical separation.

%%%%%%%%%%%%%%%%%%%%%%%%%%%%%%%%%%%%%%%%%%%%%%%%%%%%%%%%%%%%%%%%%%%%%%%%%%%%%%%%%%%%%%%%%%%%%%%%%%%%%%%%%%%%%%%%%%%%%%%%%%%%%%%%%%%%%%%%%%%%%
\section{Computational Characteristics: Robot Control}
\label{sec:control}
So far in this article, we have discussed the \textit{perceptible} aspects of an HRI system, which describe `who is present in the system and who is interacting with whom'.  Now, we look into computational aspects of the system, specifically how system designers can choose to influence/control the behavior of different agents in the system.  
In an HRI system, behavior and actions of robots are controlled directly, either governed by an optimization-based action-policy or via predefined/rule-based methods, or a combination of both.  On the other hand, human behavior can only be influenced indirectly using robots---via robots' actions, their interactions with the humans, other robots or the environment, or by explicit communication.  
Some systems make use of human behavioral models to decide robots' actions while others take a model-free approach.

In this article, our main focus is on multi-agent systems instead of the particulars of a study.  Therefore, even though both human behavior and robot control are relevant attributes under the aspect of computational characteristics, in this section we primarily discuss robot control, and only include a discussion of human behavior where required.
Specifically, we look into the two principal types of robot control implemented in multi-agent HRI systems: 1) Optimization-based control, and 2) Pre-defined and Rule-based control.  We also discuss the differences that the presence of multiple agents brings to the system.  

% \subsection{Numerical Implementation}
\subsection{Optimization-based control}
Optimization-based control refers to a framework of computing robot actions to optimize one or more performance-defining parameters established for the system.  
The performance parameters often represent factors like time of task completion \citep{hari2020approximation}, cost incurred (resources spent) \citep{dahiya2022scalable} and reward earned (value produced) \citep{swamy2020scaled}.  
In the multi-agent HRI literature, to pose the mathematical optimization problem, we find examples of systems being modeled in the form of time-series \citep{wang2014human}, outcome probabilities \citep{dahiya2022scalable, sellner2006coordinated} or Dynamic Bayesian Network \citep{fooladi2020bayesian}.  Machine-learning and other data-driven methods are also some of the tools used in optimization-based control, e.g., \citet{nam2019models, swamy2020scaled, li2021influencing}.  
% Optimization-based control often involves some form of mathematical interpretation of the system.  
Such system models are often motivated by literature from different areas of behavioral study such as psychology, economics and social sciences.  For example, \citet{shannon2016adaptive} present the Pew model from psychology, \citet{swamy2020scaled} make use of the Luce Choice model from economics, and \citet{bera2018socially} use entitativity related psychology research in their system.

Optimization-based control is seen in studies where the system behaviour can be modelled reliably using existing theories, or where researchers are trying to validate a new approach for the same.  These systems also require that there exist quantifiable and measurable parameters that can be used as optimizing metrics (such as task completing time, error rate, etc.).
When implementing such robot control in multi-agent systems, there are a few considerations to handle.
When multiple agents are present in the system, the required information access and the possibility of interactions may increase exponentially with the number of agents.  This problem has motivated a whole segment of research on the development of computationally efficient decision-making and control techniques for multi-agent HRI systems.  Among other applications, this research is seen in systems enabling a robot influence a team of multiple humans \citep{li2021influencing}, enabling multiple robots safely navigate among other robots and humans \citep{bajcsy2019scalable}, predicting human behavior while supervising multiple heterogeneous unmanned vehicles \citep{boussemart2011predictive}, and finding task allocation and sequencing for multiple robots travelling to collaborate with humans \citep{hari2020approximation}.  
When dealing with a large number of robots, human users' ability to maintain awareness of the system's state might be insufficient \citep{olsen2004fan, chien2013imperfect} and thus the users may benefit from a decision support system (DSS).  Applications of such DSSs are seen in systems enabling a human operator to assist multiple remote robots \citep{swamy2020scaled}, and in allocating operators to multiple navigating robots \citep{dahiya2022scalable, rosenfeld2017intelligent, malvankar2015optimal}.  There are also several studies on human supervision of fleets or swarms of remote robots.  \citet{lewis2013human} presents a review of systems enabling such supervision under the construct of command complexity, while \citet{kolling2015human} review research on human control of robot swarms.

\subsection{Predefined and rule-based control}
The predefined and rule-based control methods provide a convenient way of implementing robot control in an HRI system without the use of computational models or data.  This includes techniques like Wizard of Oz, expert knowledge-based \textit{if...then} rules or pre-specified sequences of actions.  These methods help simplifying the robot's decision-making and allow the researchers to focus on other aspects of Human-Robot Interaction that the system is designed to investigate, e.g., the outcome of a controlled interaction \citep{tan2019one, yang2012effect}.   

%%%%%%%%%%%%%%%%%%%%%%%%%%%%%%%%%%%%%%%%%%%%%%%%%%%%%%%%%%%%%%%%%%%%%%%%%%
In some systems, direct use of an optimization-based control is not possible (e.g., due to lack of a numerical model), and a subjective interpretation of system events is required based on expert knowledge.  Such systems often have robot action policies implemented as \textit{if...then} rules instead of numerically computed conditions.  For example, \citet{correia2018group} present an algorithm to generate robot emotions given system events, based on knowledge from psychology.   
Rule-based control is more commonly seen in studies that deal with subjective metrics of outcome (e.g., human perception of robots, workload, etc.). 
It is to be noted that the optimization-based and rule-based control methods are not mutually exclusive. It is possible to implement a combination of an optimizing policy with an expert knowledge-based model.  For example, in the system shown in Fig.~\ref{fig:interaction_models_images}(b), the robot's behavior is decided by a hybrid controller that combines manually-encoded behavioral rules and a machine learning-based mapping function \citet{alves2019empathic}.  

While the use of a rule-based control can enable a system to decide the robot's behavior based on some feedback from the environment, it is also possible to pre-specify robot behavior without any decision-making component.  This robot control method facilitates studies to manipulate different test conditions where feedback from the environment is not required.  Such implementation is commonly seen in studies which investigate the effects of specific robot's  behaviors on human users, e.g., users' perception of robots \citep{fraune2017threatening}, knowledge acquisition \citep{fernandez2020analysing}, group emotions \citep{bera2018classifying} and perceived legibility \citep{faria2021understanding}.  Referring back to the interaction graphs defined in Section~\ref{sec:general_classification}, this type of robot control results in unidirectional edges from robots to humans.

Under predefined and rule-based control methods, the Wizard of Oz (WoZ) is a common technique of implementing human-assisted robot control in multi-agent (mainly multi-human) systems.  In this technique, a hidden human `Wizard' (teleoperator) controls some or all of the robots' actions, speech, gestures and behavior, unknown to other human users in the system \citep{riek2012wizard}.  
Depending on the level of robot autonomy, the Wizard can be used to replace certain parts of the robots' perception or cognitive capabilities (e.g., \citet{johansson2013head, shiomi2009field}), thus overcoming the robots' limitations.
It is also possible to have a mixed-initiative approach where either the robots or the human user can take control of the robots' actions (e.g., \citet{jiang2015mixed, khasawneh2019human, wang2014human}).

\textbf{Note on Group/Individual behavioral parameters:}
Regardless of the type, robot control in multi-agent systems can be implemented in two ways: either based on parameters of individual agents or based on the group as a whole.  Taking the parameter of trust as an example in a system with a single human and multiple robots, one can either plan robots' actions by considering trust of the human in each individual robot \citep{wang2018trust}, or one can consider the human's trust in the whole robot team \citep{liu2019trust}.  Other examples of robot control based on individual parameters can be seen in studies with parameters like engagement \citep{leite2016autonomous} and attention \citep{yang2012effect}.  Such individual modelling provides a simple method to expand dyadic HRI research to the multi-agent setting, and to test any differences between the two.
Group-based robot control has its own benefits.  Often, the humans in the system are not working independently of other agents. So, one may find it useful to decide robot control as a function of group-based parameters, something not possible in dyadic systems. Such control is particularly useful in systems where agents act as a coordinating team, or in applications where performance or behavior of the whole group is central.
For example, a robot can express group-based emotions to increase its likability among human teammates \citep{correia2018group}, or use audio features to affect group entitativity \citep{savery2021emotion}.  Figure~\ref{fig:team_size}(a) shows a system where the robot determines the best way of approaching a group of humans by extracting social cues shown by the group \citep{tseng2016service}.

\subsection{Key Observations}
In this section, we looked at two types of robot control, one that aims to optimize a performance parameter and another that is defined using expert knowledge.  
% \blu{Technical details of advantages and drawbacks of different models just discussed and see if we can relate then to the previous sections.}

Regarding the implementation of robot control, the Wizard of Oz technique is still the most widely used control method for studies designed to evaluate a human's response to a particular type of interactions with a single robot (or a limited number of robots).  The prevalence of the Wizard of Oz technique is an indicator that a majority of the research in multi-agent Human-Robot Interaction pertains to investigating human behavior in such systems.    

Optimization-based robot control has mainly been a part of computational research in HRI and has enabled the development of efficient algorithms for large systems.  Even though we find a good amount of papers implementing such optimizing control policies on real systems, most of the work in computational research in multi-agent HRI is still seen in theoretical studies.  This is expected, as real world implementations and testing of large systems require significantly more resources. However, we hope that as robotic systems become more accessible, the computational research in multi-agent HRI will gain traction with real systems.  

We also discussed how robot control in multi-agent systems can be based on parameters of either individual agents or of the group as a whole.  While individual agent-based control enables the extension of dyadic research to the multi-agent case, group-based control allows to study and influence behavior of the whole group without requiring exact specifications for individual agents.  Group-based control has also enabled more natural robot operation when interacting with a group of humans.  In the domain of social robots, studies are becoming popular which try to improve understanding of behavioral parameters of the team/group as a whole.  Future work in this area is essential in normalizing robot presence in human society.  

\blu{\section{Interactions between the Core Aspects and Attributes}
\label{sec:interationsCoreAspects}

Looking at the multi-agent HRI literature under the three core aspects in Sections 3-5, one can observe that system attributes chosen by researchers under one core aspect are somewhat influenced by attributes under other aspects.  To understand the interaction between these core aspects, we consider a particular set of attributes under each core aspect as the ``configuration'' of the system.  For example, the team structure is defined by its size and composition.  The categorizations under different core aspects are chosen such that the configuration of each core aspect can be set separately while designing the system, i.e, one can choose any attribute from each column of the table in Figure~2 to build a multi-agent HRI system\footnote{With the exception that having a many-to-many interaction model requires both multiple humans and multiple robots.}.  This means, in a viable system, we can have a system with any of the three configurations for team size, with either homogeneous or heterogeneous agents, following one-to-one or one-to-many interaction models, using either proximate or remote modes of communication, and with the robot(s) in the system following either optimization-based or rule-based control.

However, the configuration of one core aspect can influence how another core aspect of the system should ideally be configured.  For example, if the team size is large, then a one-to-many communication model may be more efficient, especially in co-located settings, e.g., \citet{kim2020user}.  
% A large number of robots in the system make the use of one-to-one interaction model less effective.
It is also common that systems with a large number of robots have a group command node to ensure coordination among the robots, as it may be difficult for the human(s) to communicate with all the robots in the system.  This interaction model and robot control is therefore common in Human-Swarm Interaction \citep{kolling2015human}.

Similarly, the Wizard of Oz control technique is prevalent in systems with a limited number of robots because of its ease of implementation \citep{riek2012wizard}.  However, this technique may limit the scalability of such systems and restrict the researchers to use homogeneous or a limited number of robots.  This highlights the need of developing efficient tools for implementing robot decision-making in order to facilitate HRI research in large systems. In addition, a system requiring a human Wizard to closely monitor the agent(s) can be a cognitively demanding and time consuming task. % This is a necessity to enable advanced real-world HRI systems.
The choice of configuration is also influenced by the target application and objective of the study.  As seen in Section~\ref{sec:interaction_style}, systems intended for social settings are more likely to have a one-to-many interaction model.  When a human operator is required to supervise multiple robots, a simple screen-based model is a popular choice of system designers.

Having said that, we would like to emphasize that research in multi-agent HRI is still not mature enough to provide a guide to an \textit{ideal} configuration for a system, or to even determine if such an ideal configuration exists.  
For instance, even though screen-based interfaces are prevalent in systems with multiple robots so far, researchers have started exploring VR and AR-based interfaces as potentially more immersive and better options.
As the literature in this field grows, in future one might be able to tell whether these existing configurations can be considered as guidelines for designing multi-agent HRI systems, or as opportunities to explore new configurations.}

\section{Challenges and Direction for Future Work}
\label{sec:directions}
In this section, we discuss the main challenges faced in the research of multi-agent HRI systems.  We also note some of the gaps in the existing literature and identify avenues for future research which are important to fulfil the vision of a society where robots operate safely and seamlessly among humans.  To do this, we have divided this section into two parts: discussing agent-related factors and task-related factors.

\subsection{Agent-related Factors}

Throughout this survey, we saw that human behavior in an HRI system is affected by the robots' actions, the interactions they have with other agents and the mere presence of other agents in the environment.  Understanding how human behavior changes/adapts as a function of all these parameters can facilitate achieving the system's goal, whether it is to maximize a performance metric or to elicit a particular interaction outcome.
Having multiple agents in the system not only expands the possibilities of such behavioral influence but also brings some additional challenges when compared to the dyadic systems.  
Below, we discusses the challenges and prospects of research focused on agent-specific factors: both robot-related factors like physical appearance, level of autonomy and embodiment, and human-related factors like trust, situational awareness and workload: 

\begin{enumerate}[wide]

\item \textbf{Human behavior in groups}:  Our understanding of how humans behave in a group of multiple robots and humans is still limited, and thus the systems that make use of this understanding are also not commonly seen in the HRI literature. 
\blu{Therefore, HRI research will benefit from studies that investigate how human behavior changes due to the addition of other agents in the environment.  Similarly, it is also interesting to explore if the same robot actions can result in different outcomes based on the number of agents in the system.  More importantly, it is imperative that we understand how to accurately model human-robot interactions in a group/team setting.
We expect to see significant advances in such HRI systems resulting from multidisciplinary research between the  robotics research community and those who study human interactions: e.g. sociologists, psychologists and cognitive scientists.}
In a recent work, \citet{abrams2020c} present a framework to understand group dynamics in HRI based on ingroup identification, cohesion and entitativity concepts from social sciences.  However, we are yet to see implementation of such group-based behavioral concepts in fully functioning multi-agent HRI systems.

\item \textbf{Workload and Situational Awareness}: In the literature, we find ample research on how robot control can affect parameters like perceived human workload and situational awareness.  However, most of this research (in multi-agent HRI) is done in teams consisting of a single human working with multiple robots \citep{wong2017workload}.  There is a clear lack of research on how these human-related parameters are affected when working in a mixed team of multiple humans and robots.  \citet{gombolay2015decision} studied the effects of shared decision-making authority in mixed human-robot teams (a human subject working with a robot and another human).  More research is required to investigate how the presence of other humans in different roles and with different authorities affect each human user and how these effects change with an increasing number of agents in the system.

\item \textbf{Level of Autonomy}: It is well studied how the level of robot autonomy in the system can affect factors like system performance, human trust and workload \citep{beer2014toward}.  
% \citet{beer2014toward} present a taxonomy of levels of robot autonomy for HRI based on robot's sensing, planning and acting capabilities.  
When multiple robots are present, an HRI system may include robots with varied levels of autonomy, and it might not be possible to assign a single autonomy level to the whole system.  In addition, different agents can exhibit different roles, performance and identities.  The implication is that the relationship between a robot parameter and its influence on the HRI becomes much more complicated to study, especially for the human-related and interaction-related variables (see \citet{beer2014toward}).  For example, human trust has been investigated in a multi-agent setting in several studies, e.g. \citep{williams2021deconstructed, wang2018trust}.  However, there is still work to be done in understanding how trust is affected when robots differ in levels of autonomy, and exhibit varied performance and failure rates.  Notably, we do find some recent works towards this direction \citep{reig2021flailing, liu2019trust} and expect to see more research done on this topic.

\item \textbf{Human Predictability}: Having multiple humans brings a significant increase in the complexity of robot control and challenges faced.  A higher number of humans in the system often results in a decrease in system observability and predictability of its next states.  Since all human users are different from one another, the robots in the system may need to adapt their actions according to each human and remember those differences and individual interaction styles.  Developing multi-agent systems where the robot decision-making is aware of individual human teammates in relation to the whole group, can help make the human behavior more controllable and predictable \citep{martins2017bum}.

\blu{\item \textbf{Partner Selection}: In a proximate/co-located setting, it may also be crucial for the robots to dynamically choose their interaction partner(s), especially for a large team size.  In the literature we find limited studies on partner selection in human-robot teams, investigating how humans choose their robotic partners \citep{correia2019choose} and how robots can choose their human partners \citep{hasnain2012synchrony}.  In addition to partner selection, it may also be important to assess and improve the strength of interaction between the interacting partners \citep{lemaignan2016real}.  Advancing this research will help making robots' interactions with human teammates more seamless and natural, particularly in social settings.}

\item \textbf{Conflict Resolution}: With increases in the amount of computational research being conducted in the development of control policies in a multi-agent setting, we expect to see an increase in real-world implementations of multi-robot -- multi-human systems in the near future. However, this requires us to overcome challenges of establishing viable interaction models among all the agents and resolving any conflicts in roles and authority that may arise \citep{patel2020improving}.  Moreover, in systems where a high degree of coordination is required among robots, human performance suffers drastically with an increasing number of robots that are not coordinating autonomously \citep{lewis2013human}.

\end{enumerate}

\subsection{Task-related Factors}
In this part, we discuss the challenges faced on the implementation front of multi-agent HRI research.  These challenges can be specific to particular tasks or to the choice of implementation of a particular system:

\begin{enumerate}[wide]

\item \textbf{Safety}: In tasks where humans and robots share the environment, safety of human users becomes a critical requirement.  Most of the existing work on safety of robot navigation in a shared environment propose solution frameworks built around avoiding the humans, e.g., \citet{sathyamoorthy2020frozone, aroor2018online}.  There is a need to expand the work on safe robot operation to include systems where humans and robots are working collaboratively on a task, e.g., handling the same objects, sharing workspaces (and not just the environment)--- like in physical-HRI.  This requirement forces the robots to operate safely while working with the humans instead of trying to avoid them.

\item \textbf{Scalability}: In tasks where agents' decisions have to be computed online or where the control policy depends on a large number of parameters, one important consideration is the scalability, i.e., how well the robot control and decision-making technique scale with an increasing number of humans and robots in the system.  In the literature, we find a number of studies that address the issue of scalability by proposing efficient algorithms based on heuristics or approximations (e.g., \citep{dahiya2022scalable, hari2020approximation}).  However, the existing research on this topic is mostly done under a theoretical setup, and we do not see implementations of these proposed solutions in real systems with actual robots and humans (though there are a limited number of studies that conduct testing on real systems, e.g., \citep{rosenfeld2017intelligent}).  This is important because with real-world implementations of HRI systems, one also needs to consider a multitude of practical factors, and many assumptions initially made might no longer be valid and/or applicable.

\item \textbf{Transparency}: In HRI, transparency is another crucial area of research that includes the topics of motion legibility, intention recognition and decision explainability, along with several other technical, ethical and legal aspects \citep{felzmann2019robots}.  Transparency aims to provide humans with a better understanding of robots' decisions and foster human trust in the system.  An increase in transparency is shown to improve task performance in multi-agent (especially multi-robot) systems \citep{mercado2016intelligent}.  However, this area is still under-explored and research has mostly been tested in situations where the teams consist of either a single robot or only homogeneous robots (see Table~2 in \citep{felzmann2019robots}).  Although transparency has been studied in dyadic systems, it is not well known how those results translate to multi-agent systems.  Exceptions are recent related work on legibility of multi-robot teams \citep{kim2021generating, capelli2019communication} and the use of a situational awareness-based transparency framework to improve team effectiveness \citep{chen2018situation}.  We believe that more research in this area will benefit mixed teams of multiple humans and robots, particularly in direct interaction scenarios.

\end{enumerate}

\section{Conclusion}
\label{sec:conclusion}
In this article, we surveyed Human-Robot Interaction (HRI) literature for systems comprising multiple agents, be it humans or robots.  Future robots are envisioned to operate in the same environment as humans, and interact and collaborate as equal partners.  To fulfil this vision, it is crucial to promote research in multi-agent HRI systems.  Such multi-agent systems have started to gain popularity among researchers, and are enabling more realistic and practical scenarios in which humans and robots interact.  These systems do come with their own challenges, on top of the challenges that dyadic human-robot systems have, and should be considered in the development of future human-robot systems.

Through this survey, we provide the readers with a way to compare the diverse research works that exist in the field of multi-agent Human-Robot Interaction.  This survey has helped us recognize key trends in the existing research, and discuss the challenges that must be overcome.
Understanding the dynamics of Human-Robot Interaction in a multi-agent setting, and development of modelling and control strategies for such systems are crucial steps towards fulfilling the vision of the society where humans and robots exist together and interact seamlessly with each other.  Setting up multi-agent HRI systems presents a number of challenges related to safety, scalability, coordination and information transparency, which we need to address in order to progress towards this vision.  Understanding human behavior in a group setting among heterogeneous agents (both robots and humans) is important to properly model future HRI systems.  Establishing viable interaction models is another key requirement for advanced multi-agent HRI systems, especially the ones enabling interactions among multiple humans and multiple robots simultaneously.

%%%%%%%%%%%%%%%%%%%%%%%%%%%%%%%%%%%%%%%%%%%%%%%%%%%%%%%%%%%%%%%%%%%%%%%%%%%%%%%%
%\bibliographystyle{elsarticle-harv}
%\bibliography{abhinav_biblio}

%%%%%%%%%%%%%%%%%%%%%%%%%%%%%%%%%%%%%%%%%%%%%%%%%%%%%%%%%%%%%%%%%%%%%%%%%%%%%%%%%%%
\end{document}